\documentclass[10pt,twocolumn,letterpaper]{article}

\usepackage{cvpr}              % To produce the CAMERA-READY version
\usepackage{multirow}
\usepackage{booktabs}
\usepackage{graphicx}
\usepackage{multirow}
\usepackage{booktabs}
\usepackage{array} % 这里加载 array 宏包
\usepackage{amsmath}
\usepackage{float} 
\usepackage[toc,page]{appendix} % 加载 appendix 宏包
\usepackage[accsupp]{axessibility}

\definecolor{cvprblue}{rgb}{0.21,0.49,0.74}
\usepackage[pagebackref,breaklinks,colorlinks,allcolors=cvprblue]{hyperref}

\def\paperID{3500} % *** Enter the Paper ID here
\def\confName{CVPR}
\def\confYear{2025}

\title{TAET: Two-Stage Adversarial Equalization Training on Long-Tailed Distributions}

\author{
	Wang Yu-Hang\textsuperscript{1}, Junkang Guo\textsuperscript{1}, Aolei Liu\textsuperscript{1}, Kaihao Wang\textsuperscript{1},\\ 
	Zaitong Wu\textsuperscript{1}, Zhenyu Liu\textsuperscript{1}, Wenfei Yin\textsuperscript{1}, Jian Liu\textsuperscript{1,}\thanks{Corresponding author}\\
	\textsuperscript{1}School of Computer Science and Information Engineering, Hefei University of Technology\\
	{\tt\small \{wangyuhang, 2023110577, lal, 2022218133,}\\  % 邮箱换行保持对齐
	{\tt\small 2022217463, 2023217514, wenfeiyin, jianliu\}@mail.hfut.edu.cn}
}

\begin{document}
\maketitle
\begin{abstract}
	
Adversarial robustness remains a significant challenge in deploying deep neural networks for real-world applications. While adversarial training is widely acknowledged as a promising defense strategy, most existing studies primarily focus on balanced datasets, neglecting the fact that real-world data often exhibit a long-tailed distribution, which introduces substantial challenges to robustness. In this paper, we provide an in-depth analysis of adversarial training in the context of long-tailed distributions and identify the limitations of the current state-of-the-art method, AT-BSL, in achieving robust performance under such conditions. To address these challenges, we propose a novel training framework, \textbf{TAET}, which incorporates an initial stabilization phase followed by a stratified, equalization adversarial training phase. Furthermore, prior work on long-tailed robustness has largely overlooked a crucial evaluation metric—Balanced Accuracy. To fill this gap, we introduce the concept of \textbf{Balanced Robustness}, a comprehensive metric that measures robustness specifically under long-tailed distributions. Extensive experiments demonstrate that our method outperforms existing advanced defenses, yielding significant improvements in both memory and computational efficiency. We believe this work represents a substantial step forward in tackling robustness challenges in real-world applications. Our paper code can be found at \href{https://github.com/BuhuiOK/TAET-Two-Stage-Adversarial-Equalization-Training-on-Long-Tailed-Distributions}{https://github.com/BuhuiOK/TAET-Two-Stage-Adversarial-Equalization-Training-on-Long-Tailed-Distributions}

\end{abstract}    
\section{Introduction}
\label{sec:1}
% TODO: \usepackage{graphicx} required
\setlength{\textfloatsep}{5pt plus 1.0pt minus 2.0pt}
\begin{figure}
	\centering
	\includegraphics[width=1\linewidth]{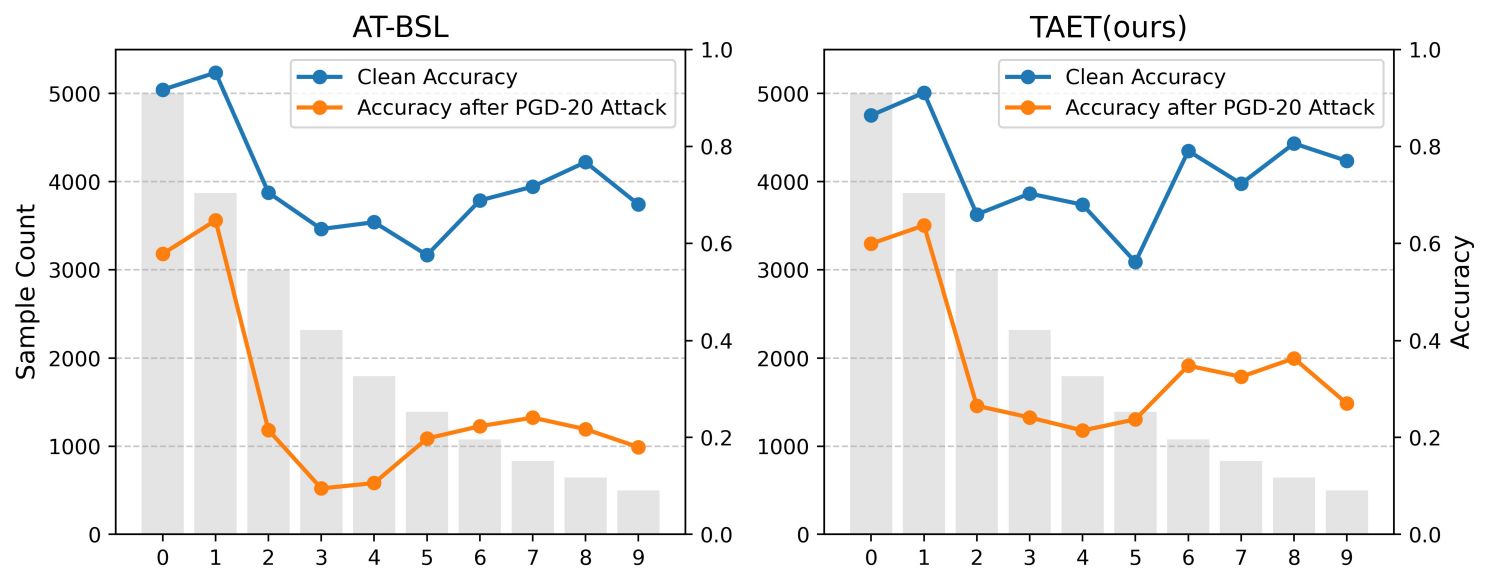}
	\includegraphics[width=0.95\linewidth]{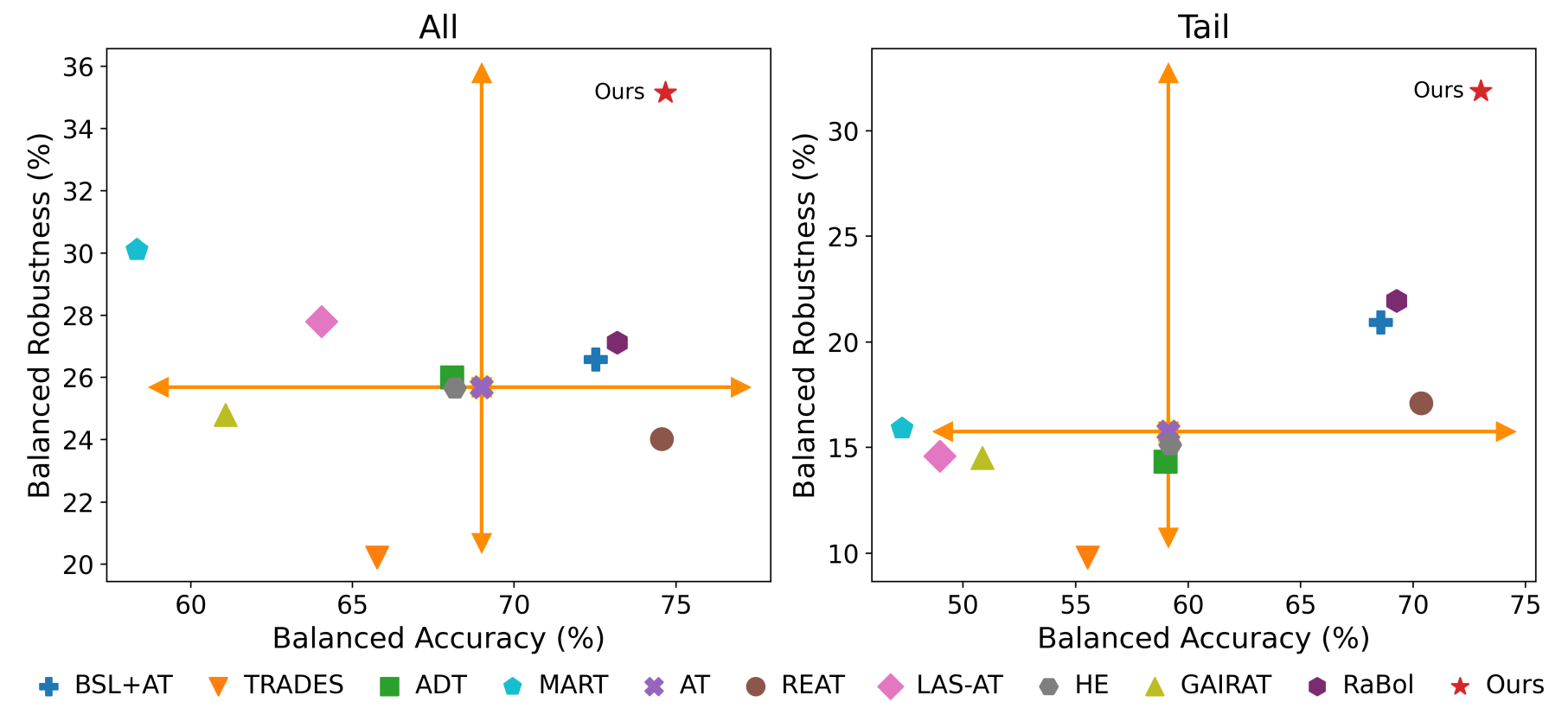}
	\caption{ \textbf{Top:} The distribution of accuracy and adversarial robustness across different classes in a long-tail distributions, with gray bars representing the sample counts of each class. AT-BSL (left) exhibits poorer performance on the tail classes (7, 8, 9) and Class 3. \textbf{Bottom:} The evaluation includes both balanced accuracy and robustness, comparing long-tail recognition methods, adversarial training, state-of-the-art defenses, and our proposed approach. The results demonstrate that our method outperforms the others in both balanced accuracy and performance on weak classes.}
	\label{fig1}
\end{figure}

In recent years, deep learning has achieved groundbreaking advancements in computer vision \cite{jinzhan}. However, deep neural networks remain highly susceptible to adversarial attacks \cite{duikang1,duikang2fgsm}, a challenge that continues to raise concerns in both academia and industry. These attacks introduce subtle perturbations to input data, leading to erroneous predictions and revealing the vulnerabilities of neural networks to malicious disturbances, which pose significant security risks to the deployment of modern computer vision models in real-world applications \cite{duikangweihai}. To address this issue, researchers have developed a variety of techniques\cite{trades,he,gairat,mart} aimed at enhancing adversarial robustness. Among these, Adversarial Training (AT) \cite{at1pgd} is widely regarded as one of the most effective strategies for improving model resilience. The central concept of AT involves integrating adversarial examples into the training process, thereby fortifying the model’s ability to recognize and counteract malicious perturbations and enhancing its generalization capability in practical scenarios, ultimately improving both reliability and security.

Adversarial training has shown strong performance on balanced datasets (e.g., MNIST, CIFAR-10 \cite{cifar10}, ImageNet \cite{imagenet}). However, real-world data often follow long-tailed distributions \cite{changweizhibiao,changweizhibiao2,changweiduikang2021}, where certain classes are overrepresented while others are underrepresented. This imbalance complicates tasks and undermines model robustness, necessitating adversarial training strategies tailored for such distributions. Long-tail learning addresses imbalanced datasets where a few head classes dominate, and many tail classes are underrepresented, creating a skewed frequency distribution \cite{changweishibie,changweishibie2,changweishibie3}. The Imbalance Ratio (IR) \cite{ir} quantifies this imbalance, with higher values indicating greater scarcity of tail class samples, exacerbating learning challenges. While long-tail learning aims to address these issues, problems like model bias toward head classes and insufficient tail class data persist \cite{changweizhibiao}. Existing robustness methods struggle with suboptimal performance on tail classes \cite{changweiduikang2021,changweiduikang2024,cahngweiduikang2023arxiv}, as shown in Fig.~\ref{fig1}. In adversarial training for long-tailed distributions, accuracy improves but adversarial robustness does not scale proportionally, leading to robustness overfitting. This phenomenon results in models becoming more vulnerable to adversarial attacks despite higher accuracy. To tackle this, we propose Hierarchical Adversarial Robust Learning (HARL), a framework designed to optimize performance on underrepresented classes and improve adversarial robustness in long-tailed distributions.

Our evaluation of robustness in long-tailed distributions reveals significant limitations in current assessment metrics \cite{changweiduikang2021, changweiduikang2024, cahngweiduikang2023arxiv}. Although balanced accuracy \cite{changweizhibiao, changweizhibiao2, changweizhibiao3, changweizhibiao4} is commonly used for long-tail recognition, traditional accuracy remains the dominant metric for evaluating long-tail robustness. To address this gap, we introduce a novel metric—\textbf{Balanced Robustness}—which, alongside balanced accuracy, provides a more comprehensive framework for evaluating robustness in long-tailed distributions. To mitigate robustness overfitting \cite{lubangguonihe} while optimizing training efficiency, we explore the potential of using cross-entropy loss during the early stages of adversarial training. We hypothesize that early-stage cross-entropy loss can improve both natural accuracy and adversarial robustness. Based on this insight, we propose a two-stage adversarial balanced training approach that optimizes both balanced robustness and accuracy, while minimizing memory usage and computational overhead, thereby improving practical applicability.

Our contributions are summarized as follows:
\begin{itemize}
	\item We analyze adversarial training under long-tailed distributions and identify a key limitation: while training accuracy improves, adversarial robustness lags behind. Our root-cause analysis informs targeted enhancements to address this gap in real-world robustness.
	\item We introduce Hierarchical Adversarial Robust Learning (HARL), a method designed to enhance robustness for underrepresented classes in long-tail distributions, achieving competitive natural accuracy and outperforming mainstream methods across multiple datasets.
	\item We propose a new evaluation metric, Balanced Robustness, which, alongside balanced accuracy, offers a more effective measure of robustness in long-tail scenarios.
	\item We present a two-stage adversarial training framework that incorporates cross-entropy loss during the early stages to improve both natural accuracy and adversarial robustness. Comparative results demonstrate that our approach optimizes memory efficiency and computational time, surpassing alternative methods.
\end{itemize}

%\begin{figure*}
%  \centering
%  \begin{subfigure}{0.68\linewidth}
%    \fbox{\rule{0pt}{2in} \rule{.9\linewidth}{0pt}}
%    \caption{An example of a subfigure.}
%    \label{fig:short-a}
%  \end{subfigure}
%  \hfill
%  \begin{subfigure}{0.28\linewidth}
%    \fbox{\rule{0pt}{2in} \rule{.9\linewidth}{0pt}}
%    \caption{Another example of a subfigure.}
%    \label{fig:short-b}
%  \end{subfigure}
%  \caption{Example of a short caption, which should be centered.}
%  \label{fig:short}
%\end{figure*}

\section{Related work}
\label{sec:Related work}

\subsection{ Long-Tailed Recognition}
\label{sec:2.1}

Long-tail distribution is a prevalent form of data imbalance in training datasets, characterized by a small number of "head" classes with abundant samples, contrasted with a majority of "tail" classes that have relatively few samples \cite{changweizhibiao}. Models trained on such imbalanced distributions typically show high confidence in predicting head classes, leading to "overconfidence" that severely hampers generalization to tail classes. This issue becomes particularly pronounced when evaluating rare categories in real-world applications, as illustrated in Fig.~\ref{fig:fig5}. Effectively addressing this challenge remains a significant research hurdle \cite{changweizongshu}.

To tackle the difficulties of long-tail recognition, various strategies have been proposed, including resampling \cite{chongcaiyang}, cost-sensitive learning \cite{cost-sensitive}, decoupled training \cite{decoupledtraining}, and classifier design \cite{classifierdesign}. These methods aim to mitigate data imbalance by altering data distributions, adjusting loss weights, and enhancing feature learning. Recent innovations, such as Class-Conditional Sharpness-Aware Minimization (CC-SAM) \cite{ccsam} and Feature Cluster Compression (FCC) \cite{fcc}, offer novel solutions to improve model generalization and recognition accuracy by refining feature learning and classification processes. However, despite advancements in these areas, few studies have specifically addressed improving adversarial robustness in long-tail distributions \cite{changweiduikang2021,changweiduikang2024,cahngweiduikang2023arxiv}, highlighting a crucial gap that warrants further investigation.

\subsection{Adversarial Robustness}
\label{sec:2.2}

To mitigate adversarial vulnerability in deep learning models, a variety of defense strategies have been proposed, including adversarial training \cite{at1pgd}, defensive distillation \cite{defensivedistillation}, ensemble methods \cite{ensemblemethods}, and data augmentation \cite{data}, with adversarial training being one of the most robust and widely adopted approaches. Techniques such as supervised learning \cite{supervisedlearning}, feature denoising \cite{featuredenoising}, and statistical filtering \cite{statisticalfiltering} further enhance model robustness by improving generalization and reducing the impact of adversarial perturbations. Due to the high computational cost associated with adversarial training, recent advancements have focused on improving its efficiency \cite{youhua}, making it more viable for deployment in real-world applications. The core principle of adversarial training involves generating the most challenging adversarial samples internally and optimizing the model's parameters based on these samples, which inherently enhances its robustness \cite{at1pgd}. The objective of adversarial training can be formalized as follows:

\begin{equation}
	\min_{\boldsymbol{\theta}} \widehat{\mathcal{R}}_{x'} = \frac{1}{|\mathcal{S}|} \sum_{(x, y) \in \mathcal{S}} \left( \max_{\boldsymbol{\delta} \in \mathcal{B}(x)} \mathcal{L}(f_{\boldsymbol{\theta}}(x + \boldsymbol{\delta}), y) \right).
\end{equation}

 \( \widehat{\mathcal{R}}_{x'} \) represents the adversarial risk for model \( f_{\boldsymbol{\theta}} \) on dataset \( \mathcal{S} \). Here, \( |\mathcal{S}| \) denotes the number of samples in \( \mathcal{S} \), with each sample consisting of an input-output pair \( (x, y) \), where \( x \) is the unperturbed input and \( y \) is the ground truth label. The perturbation \( \boldsymbol{\delta} \) is optimized within the constraint set \( \mathcal{B}(x) \), defining permissible modifications to \( x \). The formula computes the loss \( \mathcal{L}(f_{\boldsymbol{\theta}}(x + \boldsymbol{\delta}), y) \) for each perturbed sample, seeking the perturbation \( \boldsymbol{\delta} \) that maximizes this loss.

Building upon the foundation of AT, subsequent works developed advanced adversarial training techniques such as TRADES\cite{trades}, ADT\cite{adt}, MART\cite{mart}, HE\cite{he}, GAIRAT\cite{gairat}, and LAS-AT\cite{lasat}.

%-------------------------------------------------------------------------
\subsection{Robustness under Long-Tailed Distribution}
\label{sec:2.3}
Although long-tail recognition and adversarial robustness have garnered significant research interest, the challenge of adversarial robustness under long-tail distributions remains relatively underexplored. While real-world datasets often exhibit long-tail characteristics, only a limited number of studies have systematically addressed this issue \cite{changweiduikang2021,changweiduikang2024,cahngweiduikang2023arxiv}. Wu et al. \cite{changweiduikang2021} were among the first to investigate the impact of data imbalance on adversarial robustness, introducing the RoBal framework, which integrates a cosine classifier with a two-stage rebalancing strategy to enhance both natural and robust accuracy.

Building upon the RoBal framework, Yue et al. \cite{changweiduikang2024} proposed Adversarial Training with Balanced Softmax Loss (AT-BSL), a streamlined method that achieves performance comparable to RoBal while significantly reducing both training time and memory consumption. This study provides an in-depth analysis of AT-BSL, particularly focusing on the design of the Balanced Softmax Loss (BSL) in  Sec.~\ref{sec:Analysis of AT-BSL}, offering novel insights and strategies to improve adversarial robustness under long-tail distributions.

\section{Analysis of AT-BSL}
\label{sec:Analysis of AT-BSL}
In this section, we present Adversarial Training with Balanced Softmax Loss (AT-BSL)\cite{changweiduikang2024} and clarify its principles, along with defining the notation used.

\subsection{ Preliminaries}
\label{sec:3.1}
\textbf{Balanced Softmax Loss} adjusts class logits to enhance performance on long-tail datasets. It modifies the logit $z_i$ for each class, adding a term $b_y = \tau_b \log n_y$ for the target class $y$ to control class influence\cite{changweiduikang2021}. The formula is as follows:
\begin{equation}
	\begin{split}
		L_0(g(f(x)), y) &= -\log \left( \frac{e^{z_y + b_y}}{\sum_{i} e^{z_i + b_i}} \right) \\
		&= -\log \left( \frac{n_y^{\tau_b} \cdot e^{z_y}}{\sum_{i} n_i^{\tau_b} \cdot e^{z_i}} \right).
	\end{split}
\end{equation}

Here, $g(\cdot)$ denotes the linear classifier, $f(x)$ is the feature representation of input $x$, $y$ is the true class label, $z_y$ the logit for class $y$, $b_y$ the bias for class $y$, and $\tau_b$ a hyperparameter adjusting the logit range to control the softmax output distribution.

\noindent\textbf{Adversarial Training}. We previously outlined the objective of adversarial training\cite{at1pgd}. Here, we detail the steps to generate adversarial samples in the AT-BSL\cite{changweiduikang2024} framework, as shown in the following formula:
\begin{equation}
	x'^{(t+1)} = \text{Proj}_{\mathcal{B}_\infty} \left( x'^{(t)} + \alpha \cdot \text{sign} \left( \nabla_{x'^{(t)}} \mathcal{L}_{\text{CE}} (f(x'^{(t)}), y) \right) \right)
\end{equation}
Here, \( x'^{(t)} \) represents the adversarial sample at step \( t \), generated to mislead the model. \( x \) is the natural sample, and \( \mathcal{B}_{\infty}(x, \epsilon) \) defines the \( \ell_{\infty} \) norm ball around \( x \) with radius \( \epsilon \), limiting adversarial perturbations. \( \text{Proj} \) is the projection, \( \mathcal{L}_{\text{CE}} \) the cross-entropy loss, and \( \alpha \) the step size, controlling each perturbation's magnitude. \( \nabla \) is the loss gradient with respect to the current adversarial sample at step \( t \), and \( \text{sign} \) applies the gradient’s sign to ensure updates align with gradient ascent.

% TODO: \usepackage{graphicx} required
\begin{figure}
	\centering
	\includegraphics[width=1\linewidth]{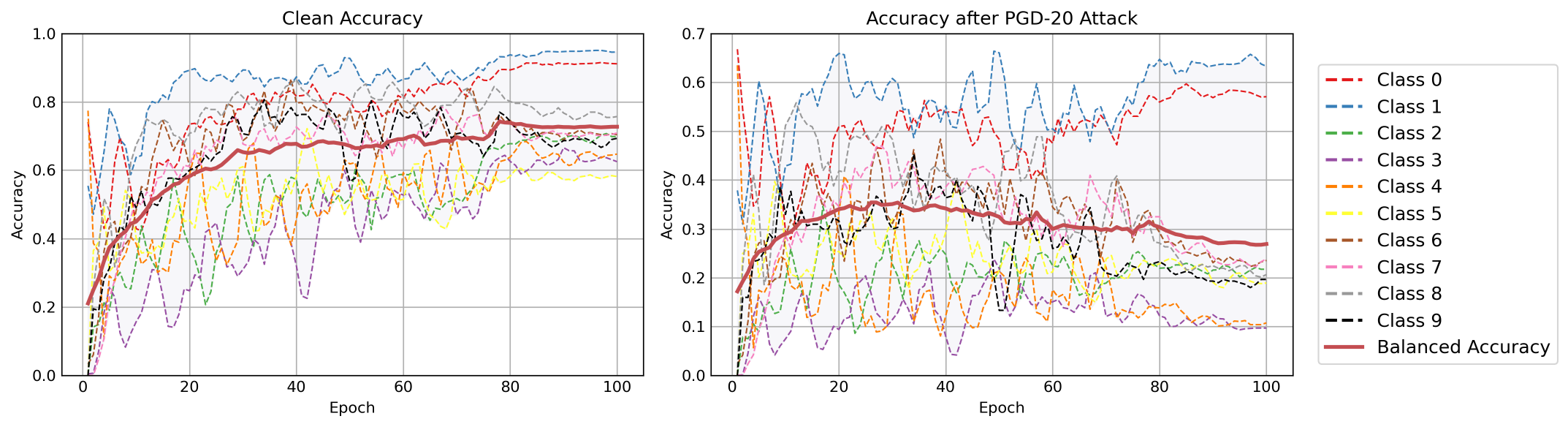}
	\includegraphics[width=1\linewidth]{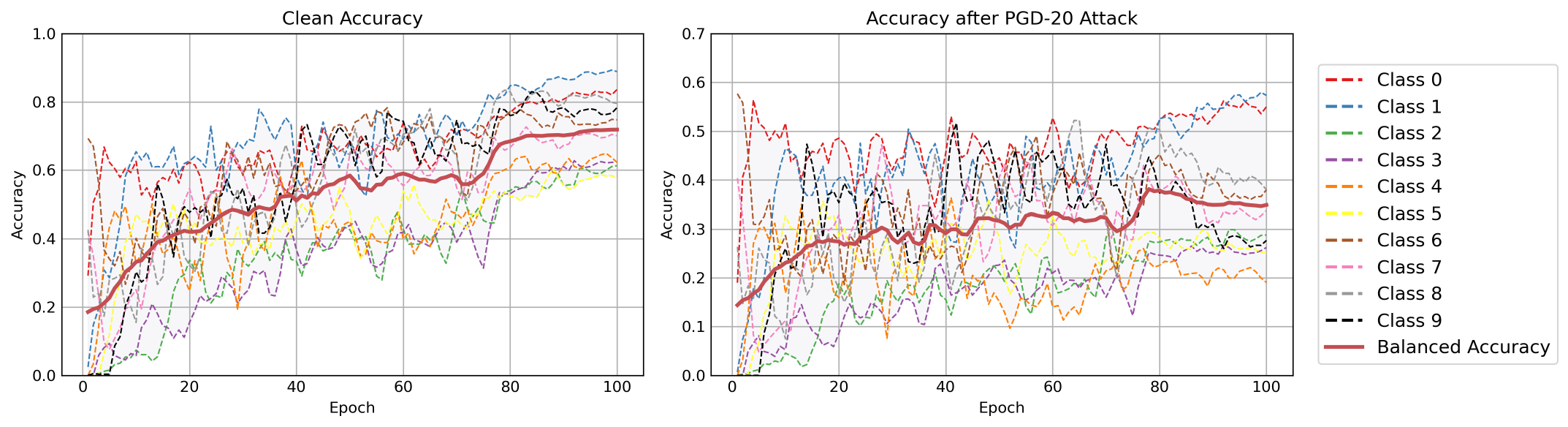}
	\caption{  \textbf{Top}: Accuracy progression during training with AT-BSL (left) and under PGD-20 attack (right). \textbf{Bottom}:Accuracy progression during training with our method (left) and under PGD-20 attack (right).}
	\label{fig:fig2}
\end{figure}

\subsection{ Limitations of AT-BSL}
\label{sec:3.2}
We conducted a thorough evaluation of AT-BSL\cite{changweiduikang2024}, with a particular focus on the variations in both accuracy and adversarial robustness across different classes throughout training. The key findings are as follows:

\noindent \textbf{The BSL method has substantial limitations in improving performance for underrepresented classes.} This method aims to enhance the performance of underrepresented classes by adjusting the output logits according to the class distribution, specifically by suppressing the logits of dominant classes in order to elevate those of minority classes. However, this straightforward approach fails to adequately improve the robustness of the weaker classes(as shown in Fig.~\ref{fig:fig2}). In contrast, our TAET method introduces a hierarchical equalization module that dynamically identifies underrepresented classes and facilitates targeted performance improvement. The experimental results indicate that our method significantly outperforms existing approaches in terms of adversarial robustness for underrepresented classes, as evidenced in Tab.~\ref{tab2}.

\begin{figure}[!t]
	\centering
	% First row: AT and TRADES
	\begin{minipage}{0.45\linewidth}
		\centering
		\includegraphics[width=\linewidth]{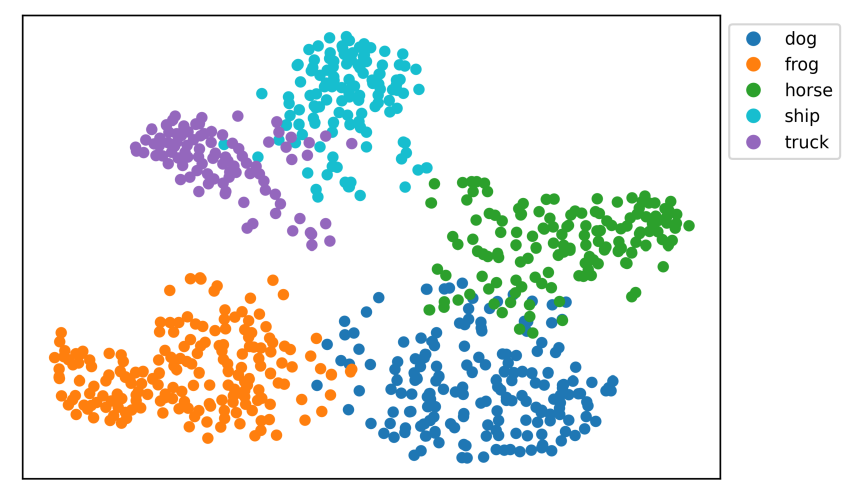}
		\vspace{-3.7ex} % Reduce space between image and subcaption
		\subcaption[AT]{\footnotesize AT}
	\end{minipage} \hspace{0.05\linewidth} % Adjust space between the images in the row
	\begin{minipage}{0.45\linewidth}
		\centering
		\includegraphics[width=\linewidth]{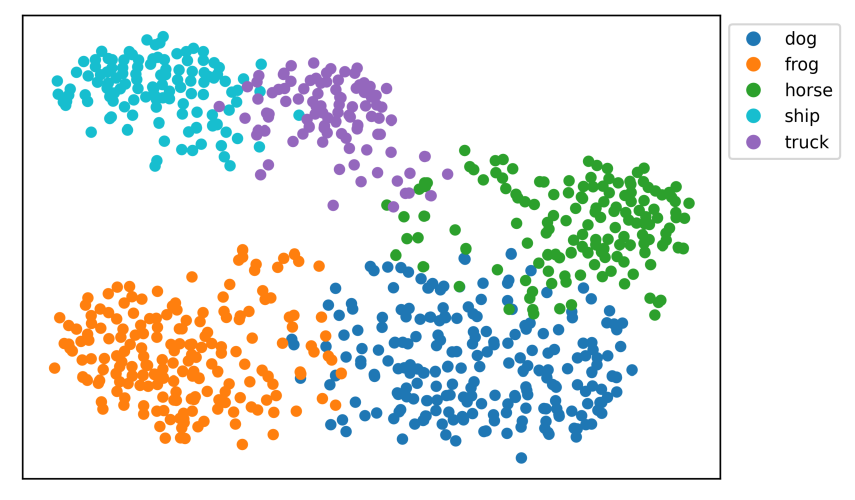}
		\vspace{-3.7ex} % Reduce space between image and subcaption
		\subcaption[TRADES]{\footnotesize TRADES}
	\end{minipage} \\[0.5ex] % Adjust the space between the rows
	
	% Second row: AT-BSL and TAET
	\begin{minipage}{0.45\linewidth}
		\centering
		\includegraphics[width=\linewidth]{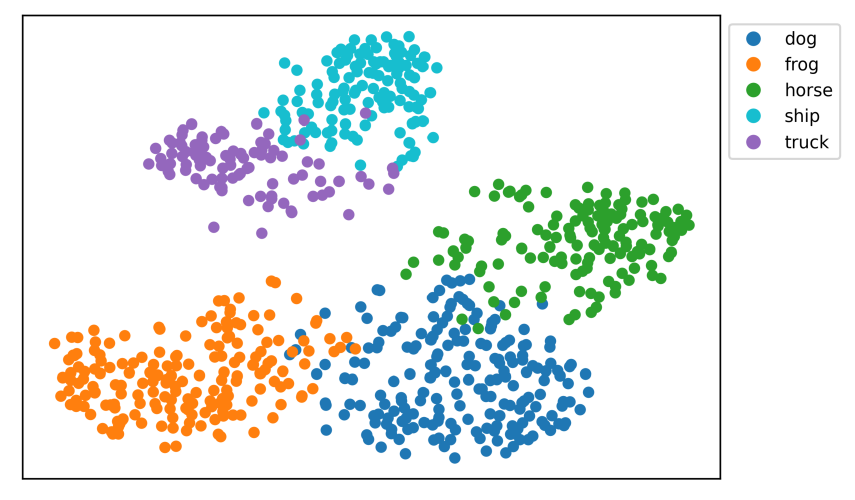}
		\vspace{-3.7ex} % Reduce space between image and subcaption
		\subcaption[AT-BSL]{\footnotesize AT-BSL}
	\end{minipage} \hspace{0.05\linewidth} % Adjust space between the images in the row
	\begin{minipage}{0.45\linewidth}
		\centering
		\includegraphics[width=\linewidth]{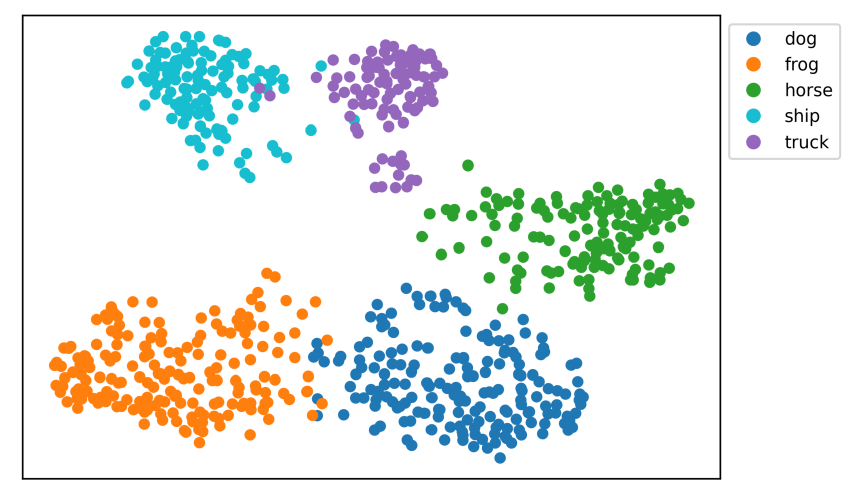}
		\vspace{-3.7ex} % Reduce space between image and subcaption
		\subcaption[TAET]{\footnotesize TAET}
	\end{minipage}
	\caption{t-SNE visualization of latent space logits extracted from (a) AT, (b) TRADES, (c) AT-BSL, and (d) our proposed TAET method on tail classes (last five classes)}
	\label{fig:fig3}
\end{figure}

\noindent \textbf{BSL is prone to robust overfitting.} Due to its fixed loss function, the model converges rapidly during training, as demonstrated in Figs.~\ref{fig:fig2} and~\ref{fig:fig3}. The robustness of BSL peaks around the 25th epoch, after which it gradually declines, particularly as the natural accuracy increases. This decline becomes more pronounced during the later training epochs (75-100), potentially causing instability when adjusting the learning rate. Although BSL improves accuracy, its adversarial robustness does not increase accordingly, and in some cases, it even decreases. As a result, it fails to achieve an optimal balance between accuracy and robustness. In contrast, our proposed method significantly improves both model robustness and accuracy, effectively reducing performance discrepancies across different classes. To address robust overfitting, we introduced a hybrid training strategy, which not only enhances model accuracy and robustness but also ensures stability when handling diverse inputs.

% TODO: \usepackage{graphicx, subcaption} required
\setlength{\textfloatsep}{5pt plus 1.0pt minus 2.0pt}

% TODO: \usepackage{graphicx} required
\begin{figure*}[!ht]
	\centering
	\includegraphics[width=1\linewidth]{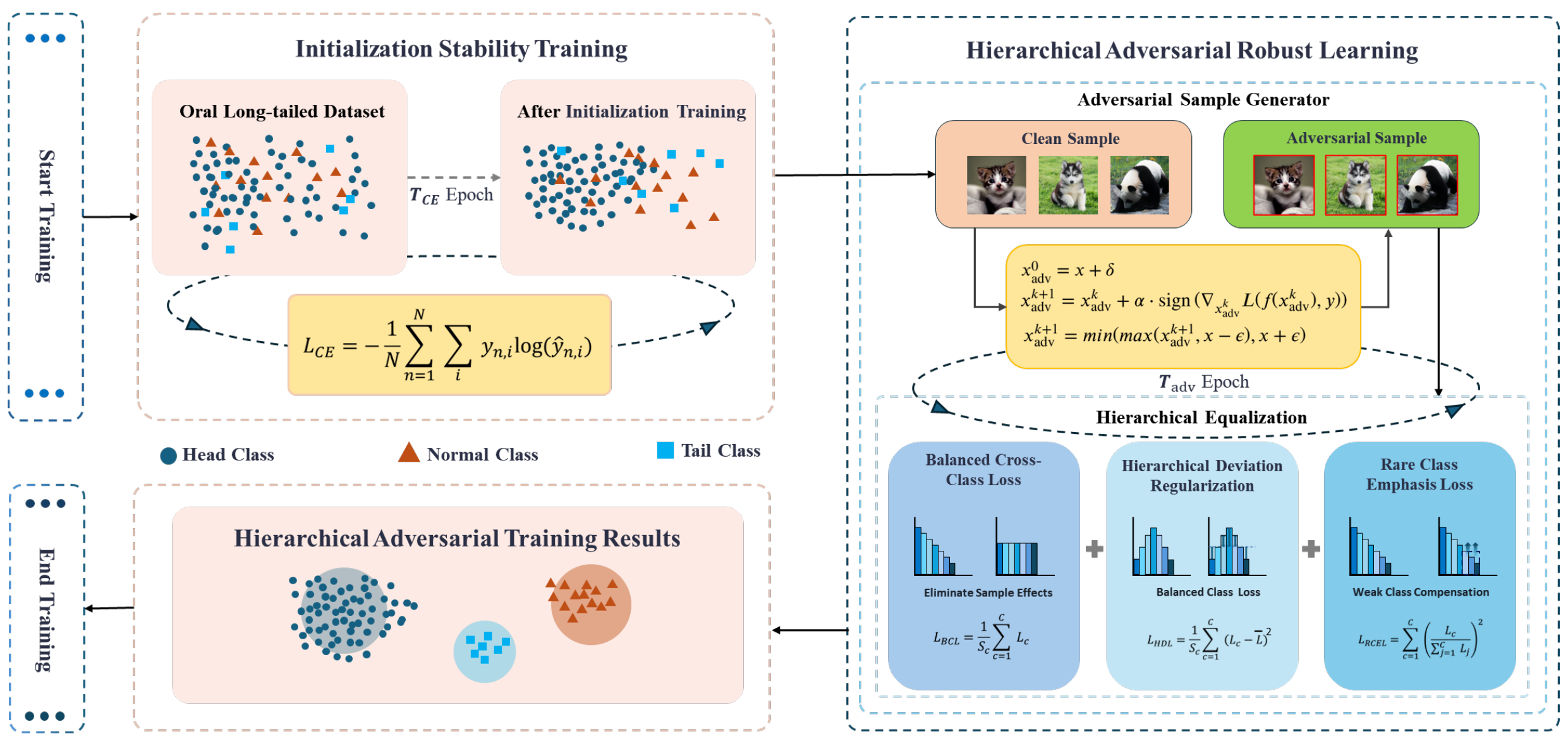}
	\caption{The TAET framework includes an Initial Stabilization Module (upper left) and the HARL module (right). The Initial Stabilization Module, based on cross-entropy loss, aims to stabilize accuracy in early training and transfers the trained model to HARL. Our HARL module consists of three components: BCL, HDL, and RCEL. A multi-step generation process creates perturbations (upper right), which are processed by normalization components (lower right).}
	\label{fig:fig4}
\end{figure*}

\section{Methodoloy}
\label{sec:methodoloy}

In Sec.~\ref{sec:Analysis of AT-BSL}, we conduct a thorough analysis of existing long-tail robustness solutions and identify two major limitations. First, current methods lack an effective mechanism for accurately identifying and compensating for the tail classes. To address this, we propose a novel adversarial equalization module, termed the Hierarchical Adversarial Robust Learning (HARL) framework, which consists of three key components, as detailed in Sec.\ref{sec:4.1}. Second, existing methods are prone to overfitting during training. To mitigate this issue, we introduce a two-stage adversarial equalization training approach, the steps of which are elaborated in Sec.~\ref{sec:4.2}.

%-------------------------------------------------------------------------
\subsection{Hierarchical Adversarial Robust Learning }
\label{sec:4.1}

Our Hierarchical Adversarial Robust Learning (HARL) framework includes an adversarial sample generation module and a hierarchical equalization component. The hierarchical equalization loss integrates Balanced Cross-Class Loss (BCL), Hierarchical Deviation Loss (HDL), and Rare Class Emphasis Loss (RCEL), working together with the adversarial sample generation module to balance accuracy and robustness. This design addresses long-tail dataset challenges, boosting minority class performance and reducing overfitting, enhancing robustness in practical applications. Unlike BSL, which uses sample count for tail class identification, HARL identifies weak classes based on mean loss during training, strengthening both tail and weak classes. Sample-count methods often miss weak classes (e.g., Class 3 in Fig.~\ref{fig1}), where large sample counts don’t guarantee good performance, a limitation our method largely overcomes.

Adversarial Sample Generation: We employ a multi-step adversarial sample generation process, which begins with random perturbations and iteratively computes the gradient of the sample with respect to the loss function. This gradient guides the direction for generating subsequent adversarial samples. A primary challenge addressed in this study is the generation of balanced adversarial samples in long-tail distributions. The transformation of adversarial samples is facilitated through hierarchical equalization loss. The learning objective for this module is as follows:
\begin{equation}
	\begin{aligned}
		\arg\min_{\boldsymbol{\theta}} &\; \mathbb{E}_{(x_0, y) \sim {D_{lt}}} \left[ \mathcal{L}_{\text{HEL}} \left( f_{\boldsymbol{\theta}}(x'), y \right) \right] \\
		\text{s.t.} &\quad \| x' - x_0 \|_{\infty} \leq \epsilon,
	\end{aligned}
\end{equation}

Here, $L_{\text{HEL}}$ represents the Hierarchical Equalization Loss\cite{hel1,hel2}, adjusting model logits for adversarial samples\cite{at1pgd}. $D_{\text{lt}}$ is the long-tail dataset, and $\epsilon$ denotes the allowable perturbation budget. The objective is to keep perturbations near the input sample $x_0$ while significantly altering the model $F$'s predictions. This aligns with prior adversarial training methods, but our approach uniquely integrates long-tail recognition with adversarial training. As shown in Fig.~\ref{fig:fig4}, we first generate adversarial samples, then apply hierarchical equalization to enhance weak classes. With a fixed perturbation budget, our goal is to minimize adversarial loss for long-tail recognition, with the loss defined as:
\begin{equation}
	\label{eq:5}
	\mathcal{L}_{\text{HEL}} = \alpha \cdot \mathcal{L}_{\text{BCL}} + \beta \cdot \mathcal{L}_{\text{HDL}} + \gamma \cdot \mathcal{L}_{\text{RCEL}}
\end{equation}
Here, $\alpha$, $\beta$, and $\gamma$ are hyperparameters for loss reweighting, tuned through validation. Each loss component is detailed below.

\noindent\textbf{Balanced Cross-Class Loss (BCL)} ensures balanced losses across classes. By reweighting, it prevents excessive focus on head classes in long-tail datasets, enabling more equal contribution to total loss. BCL achieves this by averaging class losses, enhancing classifier performance across all classes under long-tail distribution, as shown below:
\begin{equation}
	\mathcal{L}_{\text{BCL}} = \frac{1}{S_c} \sum_{c=1}^{C} \mathcal{L}_c
\end{equation}
Here, $S$ is the total number of classes, and $L_c$ represents the loss for class $c$. Averaging across all classes helps the model maintain balanced attention in a long-tail setting, avoiding bias toward head classes.

\noindent\textbf{Hierarchical Deviation Loss (HDL)} adjusts the discrepancies in inter-class loss to mitigate extreme imbalances. It quantifies the deviation of each class's loss from the mean and applies a quadratic penalty to reduce this gap, thereby improving robustness against imbalanced data distributions. This hierarchical framework allows the model to better accommodate the complexities inherent in different class distributions, as illustrated below:
\begin{equation}
	\mathcal{L}_{\text{HDL}} = \frac{1}{S_c} \sum_{c=1}^{C} (\mathcal{L}_c - \overline{\mathcal{L}})^2
\end{equation}
\noindent\textbf{Rare Class Emphasis Loss (RCEL)} addresses the challenge of long-tail distributions by focusing on rare classes, which are typically more difficult to classify. By normalizing class-wise losses and assigning higher weights to rare classes, RCEL amplifies their loss contributions. This approach encourages the model to prioritize learning from these underrepresented classes, thereby improving overall classification performance, as demonstrated below:
\begin{equation}
	\mathcal{L}_{\text{RCEL}} = \sum_{c=1}^{C} \left( \frac{\mathcal{L}_c}{\sum_{j=1}^{C} \mathcal{L}_j} \right)^2
\end{equation}

\subsection{Two-Stage Adversarial Equalization Training}
\label{sec:4.2}

In Fig.~\ref{fig:fig4}, we present our long-tail robustness framework, which initially employs a cross-entropy loss function. Our experiments show that cross-entropy loss\cite{ce} facilitates rapid convergence on unperturbed samples, ensuring early-stage accuracy. It provides stable gradient signals, enhancing robustness against adversarial attacks and mitigating robust overfitting. For a detailed experimental analysis, please refer to Sec.~\ref{sec:Experiment}.

\begin{equation}
	\vspace{-5pt} % 上方间距减少
	\mathcal{L}_{\text{CE}} = -\frac{1}{N} \sum_{n=1}^{N} \sum_{i} y_{(n,i)} \log \left( \hat{y}_{(n,i)} \right)
	\vspace{-1pt} % 上方间距减少
\end{equation}
n the later stages of training, we incorporate adversarial training along with the hierarchical equalization component to improve model performance. To highlight the contribution of hierarchical equalization within the TAET framework, we conduct an ablation study in Sec.~\ref{sec:6.4} ,evaluating the impact of each component individually. By selectively removing components, we examine how different configurations affect both accuracy and robustness, thereby identifying the optimal configuration. Fig.~\ref{fig:fig6b} illustrates the training procedure and component interactions.

\section{Revisiting Evaluation Metrics for the Long-Tail Robustness Problem}
\label{sec:5}

\noindent\textbf{Balanced Accuracy}: In long-tail robustness research, common evaluation metrics include clean accuracy and post-attack robustness. However, Balanced Accuracy (BA), an important metric for long-tail learning, is often overlooked \cite{changweizhibiao,changweizhibiao2,changweizhibiao4,changweishibie}. BA balances performance across classes by considering both true positive and true negative rates, mitigating class bias. In imbalanced datasets, traditional accuracy metrics may neglect minority classes due to the majority class's dominance. In contrast, BA ensures fair evaluation by capturing minority class performance, offering a comprehensive assessment of overall model performance. The mathematical definition of BA \cite{ba1} is as follows:

\begin{equation}
	\vspace{-5pt} % 上方间距减少
	\text{Balanced\_Accuracy} = \frac{1}{S_C} \sum_{i=1}^{C} A_i^{x_0}
	\vspace{-1pt} % 上方间距减少
\end{equation}
Where $A_i = \frac{TP_i}{TP_i + FN_i}$, $TP_i$ represents the number of true positives for class $i$, and $FN_i$ represents the number of false negatives for class $i$. The following confusion matrix pro-vides additional detail.

\noindent\textbf{Balanced Robustness}: Balanced accuracy reflects model performance across classes\cite{ba1,ba2}, mitigating bias from class imbalance, which is crucial for long-tail recognition. In long-tail robustness, we aim for consistent model performance across all classes under adversarial attacks. We extend balanced accuracy to balanced robustness, quantifying the model's average defense capability against attacks across classes. Specifically, balanced robustness, under adversarial samples, is defined as follows:\begin{equation}
		\vspace{-5pt} % 上方间距减少
	\begin{aligned}
		\text{Balance\_Robustness} &= \frac{1}{S_C} \sum_{i=1}^{C} \mathcal{R}_i^{x'} \\
		&= \frac{1}{S_C} \sum_{i=1}^{C} \frac{\text{TP}_i^{x'}}{\text{TP}_i^{x'} + \text{FN}_i^{x'}}
	\end{aligned}
		\vspace{-1pt} % 上方间距减少
\end{equation}

We introduce balanced robustness, a novel metric in the context of long-tail learning, to the best of our knowledge, marking its first application in this domain.We believe this metric can effectively measure the adversarial robustness of methods under long-tail distributions and will have extremely important applications in fields such as medicine. We advocate for the adoption of this metric as a standardized evaluation tool to promote consistency in the research and practical applications of long-tail robustness and to drive its further development.
\begin{table}[t!]
	\centering
	\setlength{\tabcolsep}{3pt} % 减少列间距
	\caption{Balanced accuracy and balanced robustness of various algorithms using ResNet-18 on CIFAR-10-LT with an imbalance ratio (IR) of 10. The best results are highlighted in \textbf{bold}.}
	\label{tab:1}
	\begin{tabular}{ccccccc}
		\toprule
		 \multirow{2}{*}{\textbf{Method}} & \multirow{2}{*}{\textbf{Clean}} & \multirow{2}{*}{\textbf{FGSM}} & \multicolumn{2}{c}{\textbf{PGD}} & \multirow{2}{*}{\textbf{CW}} & \multirow{2}{*}{\textbf{AA}} \\
		& & & \textbf{20} & \textbf{100} & & \\
		\cmidrule(r){1-7} 
		AT-BSL\cite{changweiduikang2024}      & 72.74 & 34.13 & 26.86 & 25.62 & 15.67 & 25.26 \\
		RoBal\cite{changweiduikang2021}       & 73.18 & 33.14 & 27.12 & 26.98 & 13.44 & 24.13 \\
		TRADES\cite{trades}      & 65.77 & 25.73 & 20.23 & 19.59 & 27.42 & 19.63 \\
		AT\cite{at1pgd}          & 69.00 & 32.53 & 25.69 & 24.55 & 15.22 & 24.28 \\
		MART\cite{mart}       & 58.33 & 33.67 & 30.10 & 29.36 & 48.12 & 26.04 \\
		ADT\cite{adt}         & 68.08 & 32.70 & 26.00 & 25.27 & 10.05 & 25.06 \\
		REAT\cite{cahngweiduikang2023arxiv}        & 74.56 & 31.42 & 24.02 & 22.52 & 11.07 & 22.69 \\
		LAS-AT\cite{lasat}      & 64.04 & 33.04 & 27.80 & 26.74 & 36.77 & 24.71 \\
		HE\cite{he}          & 68.18 & 32.40 & 25.66 & 24.59 & 12.94 & 24.32 \\
		GAIRAT\cite{gairat}      & 61.07 & 28.90 & 24.79 & 23.81 & 42.62 & 24.59 \\
		\cmidrule(r){1-7} 
		TAET (our) & \textbf{74.67} & \textbf{39.59} & \textbf{35.15} & \textbf{34.29} & \textbf{57.59} & \textbf{30.57} \\
		\bottomrule
	\end{tabular}
\end{table}

\setlength{\textfloatsep}{5pt plus 1.0pt minus 2.0pt}
\begin{table*}[ht]
	\centering
	\setlength{\tabcolsep}{2pt} % Reduce column spacing
	\caption{Performance of various methods on the last Tail classes (last five classes) under clean and attacked conditions using ResNet-18 on CIFAR-10-LT(IR=10). The best results are highlighted in \textbf{bold}, and the second-best results are underlined.}
	\label{tab2}
	\begin{tabular}{>{\centering\arraybackslash}p{0.15\linewidth}*{5}{>{\centering\arraybackslash}p{0.075\linewidth}}*{5}{>{\centering\arraybackslash}p{0.075\linewidth}}} 
		\toprule
		\multirow{2}{*}{\textbf{Method}} & \multicolumn{5}{c}{\textbf{Clean}} & \multicolumn{5}{c}{\textbf{Attacked (PGD-20)}} \\
		\cmidrule(lr){2-6}  \cmidrule(lr){7-11}
		& \textbf{Dog} & \textbf{Frog} & \textbf{Horse} & \textbf{Ship} & \textbf{Truck} & \textbf{Dog} & \textbf{Frog} & \textbf{Horse} & \textbf{Ship} & \textbf{Truck} \\
		\midrule
		AT-BSL\cite{changweiduikang2024}   & \underline{58.63} & 73.48 & 70.48 & 78.29 & 66 & \underline{23.02} & \underline{26.97} & 21.7 & \underline{21.7} & 19 \\
		RoBal\cite{changweiduikang2021}    & \textbf{59.13} &\underline{74.73} & 68.52 & \underline{78.96} & 65 & \textbf{23.74} & 25.78 & 20.79 & 19.5 & \underline{20} \\
		TRADES\cite{trades}   & 53.96 & 65.11 & 63.25 & 47.28 & 47 & 11.87 & 13.2 & 16.27 & 3.87 & 4 \\
		AT\cite{at1pgd}       & 51.07 & 62.32 & 63.85 & 64.34 & 54 & 14.74 & 17.67 & 19.27 & 17.05 & 10 \\
		ADT\cite{adt}      & 52.15 & 65.58 & 65.66 & 56.59 & 55 & 16.18 & 16.74 & 18.67 & 10.07 & 10 \\
		MART\cite{mart}     & 37.76 & 52.09 & 58.43 & 47.25 & 41 & 17.62 & 19.06 & \underline{25.3} & 8.52 & 9 \\
		REAT\cite{cahngweiduikang2023arxiv}     & \underline{58.63} & 73.02 & \textbf{72.89} & 78.29 & \underline{69} & 16.18 & 13.48 & 18.67 & 20.15 & 17 \\
		LAS-AT\cite{lasat}   & 50.72 & 55.34 & 56.02 & 48.83 & 34 & 19.06 & 13.02 & 20.48 & 12.4 & 8 \\
		HE\cite{he}       & 57.31 & 67.9 & 65.66 & 51.16 & 54 & 14.74 & 20 & 18.07 & 10.85 & 12 \\
		GAIRAT\cite{gairat}   & 48.56 & 58.13 & 56.62 & 48.06 & 43 & 15.82 & 17.2 & 22.28 & 6.2 & 11 \\
		\cmidrule(r){1-11}
		TAET (our)    & 56.11& \textbf{79.06} & \underline{72.28} & \textbf{80.6} & \textbf{77} & \textbf{23.74} & \textbf{34.83} & \textbf{32.53} & \textbf{36.34} & \textbf{27} \\
		\noalign{\hrule height 1pt}
	\end{tabular}
\end{table*}

\setlength{\textfloatsep}{5pt plus 1.0pt minus 2.0pt}
% TODO: \usepackage{graphicx} required
\begin{figure}
	\centering
	\includegraphics[width=1\linewidth]{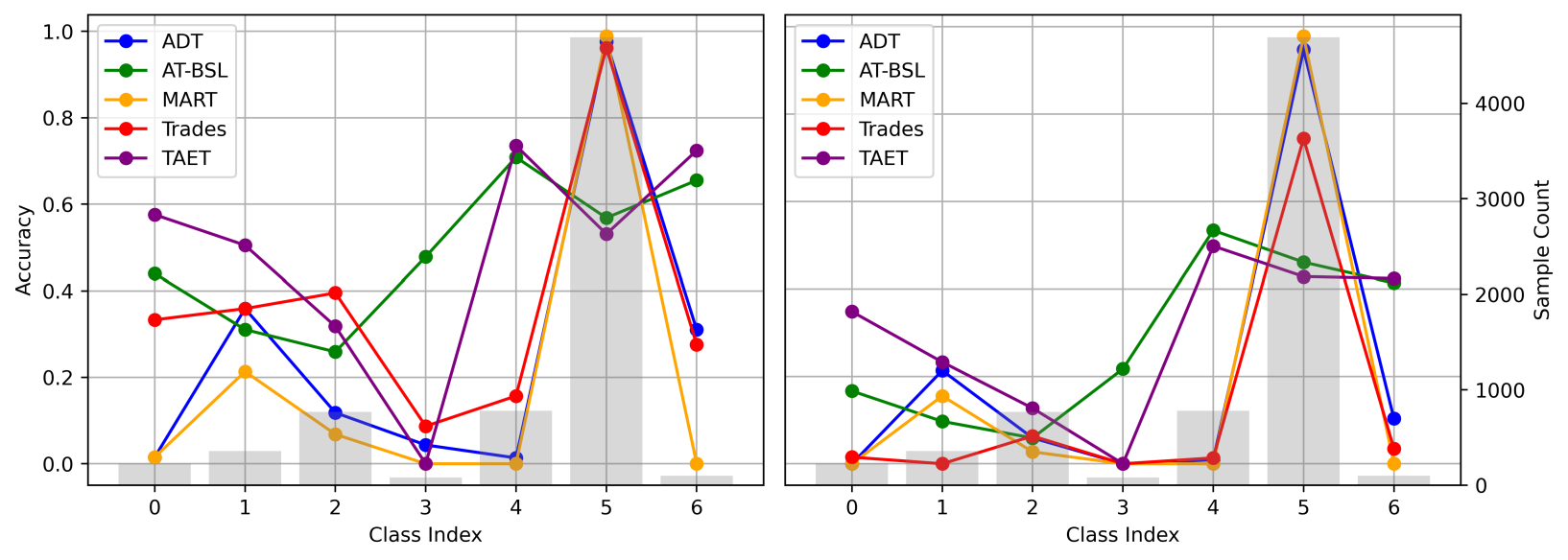}
	\caption{Results after training the VIT-B/16 model on the DermaMNIST for 100 epochs. The left side of the figure shows balanced accuracy of different methods in a natural setting, while the right side presents balanced robustness under a PGD-20 attack. The background section illustrates data distribution in DermaMNIST.}
	\label{fig:fig5}
\end{figure}

\begin{figure}[h!]
	\centering
	\begin{minipage}{0.5\linewidth}
		\centering
		\includegraphics[width=\linewidth]{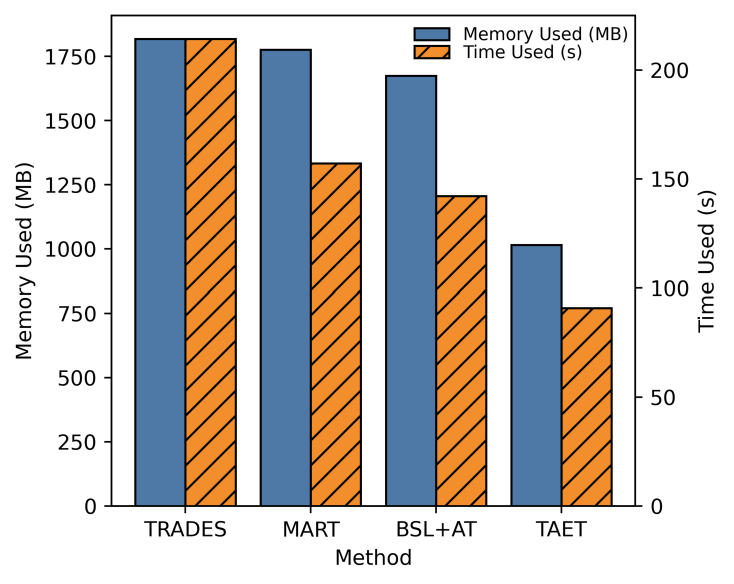}
		\subcaption{}
		\label{fig:fig6a}
	\end{minipage}
	\hfill
	\begin{minipage}{0.48\linewidth}
		\centering
		\includegraphics[width=\linewidth]{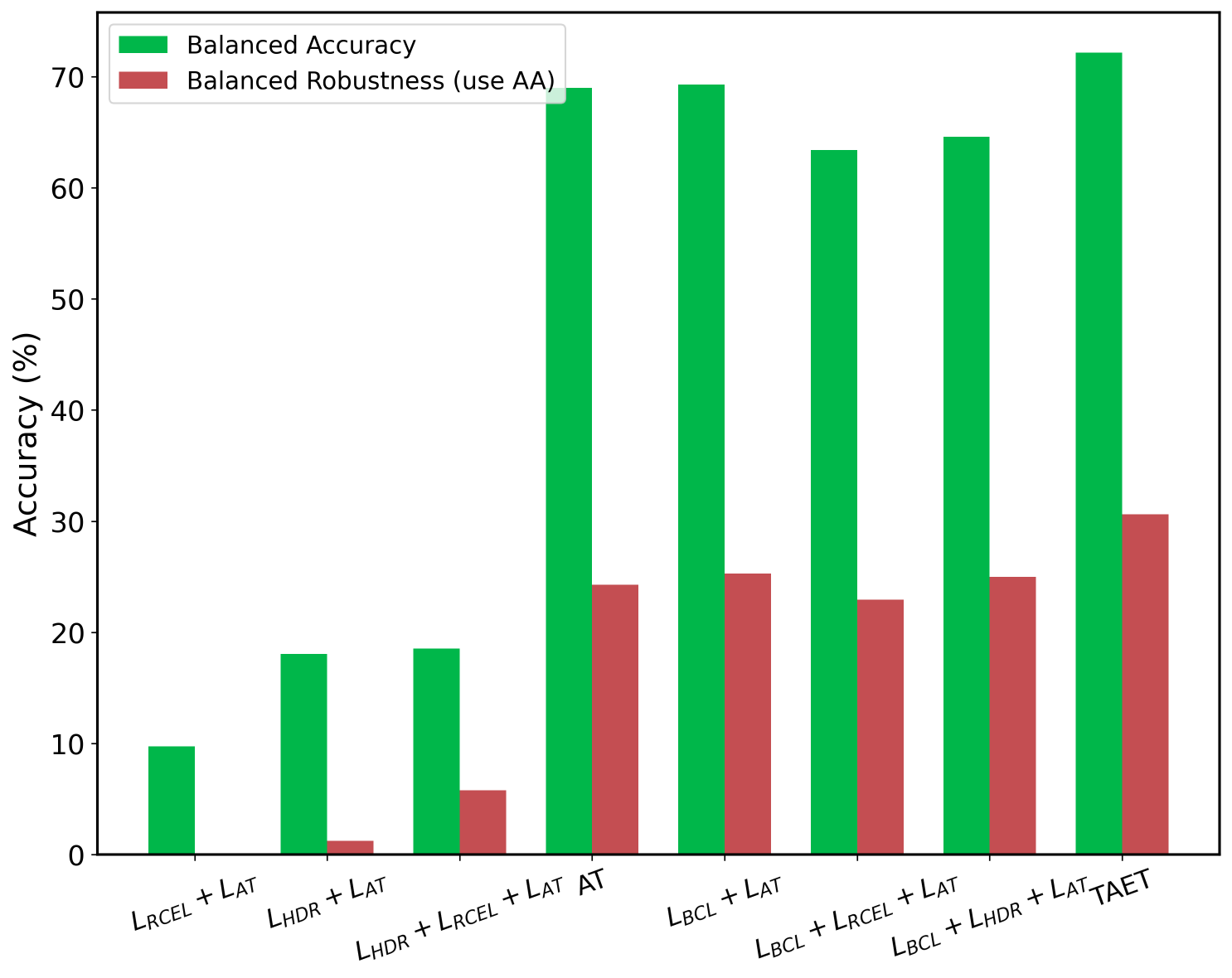}
		\subcaption{}
		\label{fig:fig6b}
	\end{minipage}
	\caption{ (a) Memory usage and time per epoch for each method; (b) Robustness and clean accuracy performance based on the effectiveness of each component in Eq. (\ref{eq:5}).}
	\label{fig:fig6}
\end{figure}

% TODO: \usepackage{graphicx} required
%\begin{figure}
%	\centering
%	\includegraphics[width=1\linewidth]{fig6}
%	\caption{ Compared to other state-of-the-art robustness methods and long-tail robustness recognition methods, our TAET approach achieves optimal robustness and clean accuracy while also minimizing memory consumption (MB) and time per epoch (seconds). The experiment was conduct-ed on hardware equipped with an NVIDIA A40 GPU, using the CIFAR-10-LT dataset with a batch size of 128.}
%	\label{fig:fig6}
%\end{figure}
%\begin{figure}
%	\centering
%	\includegraphics[width=0.5\linewidth]{fig7}
%	\caption{ Compared to other state-of-the-art robustness methods and long-tail robustness recognition methods, our TAET approach achieves optimal robustness and clean accuracy while also minimizing memory consumption (MB) and time per epoch (seconds). The experiment was conduct-ed on hardware equipped with an NVIDIA A40 GPU, using the CIFAR-10-LT dataset with a batch size of 128.}
%	\label{fig:fig7}
%\end{figure}

% 第二个表格

\vspace{-8pt} % 调整整个节的间距
\section{Experiment}
\label{sec:Experiment}
\vspace{-6pt} % 调整整个节的间距
\subsection{Settings}
\label{sec:6.1}
\vspace{-6pt} % 调整整个节的间距
 Following \cite{changweiduikang2021}, we evaluated our approach on CIFAR-10-LT and CIFAR-100-LT, and extended experiments to the MedMNIST \cite{medmnist} for real-world validation. MedMNIST consists of 12 medical image datasets across various modalities (e.g., CT, X-ray, ultrasound, OCT), supporting tasks such as multi-label and binary classification. We selected DermaMNIST, a highly imbalanced dataset from HAM10000 \cite{ham10000} with an imbalance ratio (IR) of 58.66, containing 10,015 images in 7 diagnostic categories.

The primary metrics used were \textbf{Balanced Accuracy} \cite{ba1} and \textbf{Balanced Robustness}, with dataset imbalance assessed by the imbalance ratio (IR). Model robustness was evaluated under $l_{\infty}$ bounded perturbations with a size of $8/255$ using FGSM \cite{duikang2fgsm} and PGD \cite{at1pgd} (20 and 100 steps, step size $1/255$), as well as a 100-step CW \cite{cw} attack and AutoAttack (AA) \cite{aa}, a powerful ensemble method.

\subsection{Main Results}
\label{sec:6.2}
The results in Tab.~\ref{tab:1} show that on CIFAR-10-LT, our method achieves the highest balanced accuracy and robustness on ResNet-18, with a 5.31\% improvement in robustness against AA \cite{aa} attacks compared to the AT-BSL \cite{changweiduikang2024} model. Overall, our method outperforms other methods in handling long-tail datasets.

Tab.~\ref{tab2} provides a detailed breakdown of clean accuracy and robustness in the Tail Class, demonstrating that our method significantly improves tail class performance. This is critical in real-world scenarios where tail class accuracy is low, such as in rare disease research. Our method also improves robustness for challenging classes like Class 3.

To validate our method on real-world long-tail datasets, Tab.~\ref{tab3} presents results on the DermaMNIST subset of MedMNIST \cite{medmnist}, showing optimal adversarial robustness and competitive balanced accuracy. Fig.~\ref{fig:fig5} compares clean accuracy and adversarially perturbed accuracy across classes. Our method outperforms others on tail classes such as Class 0, 1, 2, and 4. For extremely rare classes like Class 3 and Class 6, all methods show high variability. While TAET's hierarchical equalization improves tail class performance, weight adjustments for very rare classes may still be insufficient. Increasing the hyperparameter $\gamma$ in Eq. (\ref{eq:5}) could enhance these classes, though it might impact overall performance.

\vspace{-5pt} % 上方间距减少
%-------------------------------------------------------------------------
\begin{table}[!t]
	\centering
	\setlength{\tabcolsep}{4pt} % Adjust column spacing to fit single-column display
	\caption{Balanced accuracy and balanced robustness under Various Attack Methods on the MedMNIST , best results are highlighted in \textbf{bold}, and the second-best results are underlined.}
	\label{tab3}
	\begin{tabular}{cccccccc}
		\toprule
		\multirow{2}{*}{\textbf{Method}} & \multirow{2}{*}{\textbf{Clean}} & \multirow{2}{*}{\textbf{FGSM}} & \multicolumn{2}{c}{\textbf{PGD}} & \multirow{2}{*}{\textbf{CW}} & \multirow{2}{*}{\textbf{AA}} \\ 
		& & & \textbf{20} & \textbf{100} & & \\
		\cmidrule(r){1-7} 
		ADT & 26.23 & 20.07 & 19.04 & 18.93 & 23.89 & 17.81 \\
		MART & 18.37 & 17.71 & 16.59 & 16.07 & 17.61 & 15.09 \\
		TRADES & 18.89 & 16.57 & 12.48 & 11.35 & 31.18 & 10.50 \\
		AT-BSL & \textbf{48.55} & \underline{38.81} & \underline{27.85} & \underline{26.72} & \textbf{43.44} & \underline{21.04} \\
		TAET (our) & \underline{48.42} & \textbf{40.12} & \textbf{29.26} & \textbf{28.79} & \underline{43.11} & \textbf{21.74} \\
		\bottomrule
	\end{tabular}
	\vspace{-1ex} % Adjust space below the table
\end{table}

\subsection{Initial Training Epochs and Model Robustness}

\label{sec:6.3}
\begin{figure}
	\centering
	\includegraphics[width=1\linewidth]{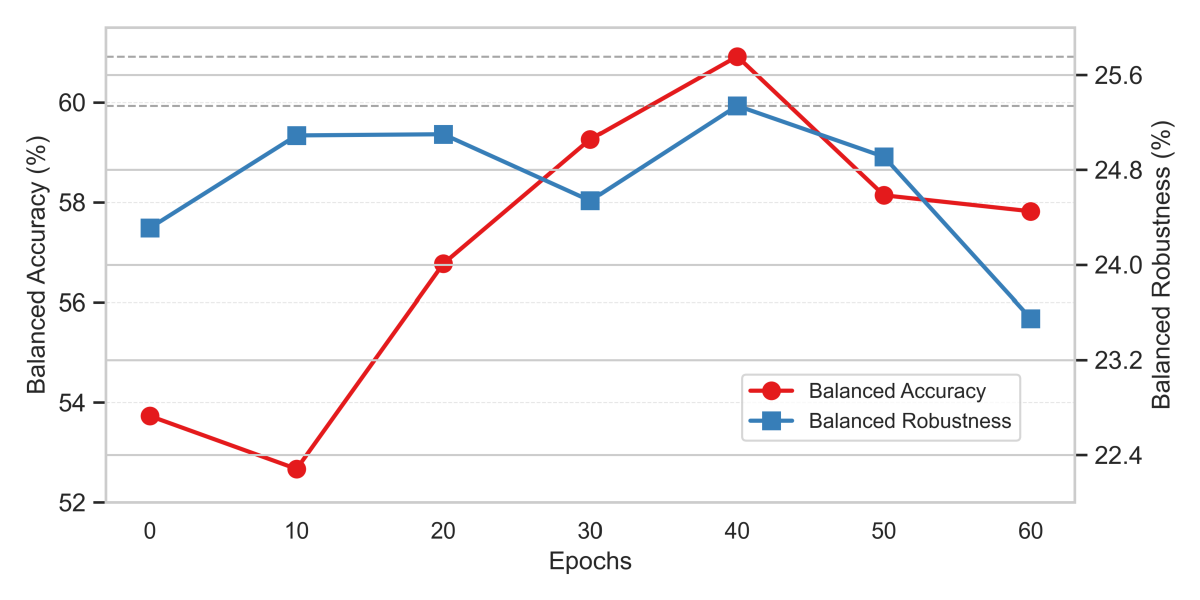}
	\caption{  Results of using different epochs of cross-entropy loss in the early training stages.}
	\label{fig:fig7}
\end{figure}

\vspace{-5pt} % 上方间距减少
To address robust overfitting\cite{lubangguonihe,lubangguonihe2}, we hypothesize that limited model accuracy may constrain robustness. Thus, improving accuracy on clean samples could also enhance robustness. Previous studies used data augmentation with some success. In our approach, we apply cross-entropy loss for a set number of epochs in the initial training phase. Experiments show that using cross-entropy loss\cite{ce}  for 40 
\begin{table}[t]
	\centering
	\setlength{\tabcolsep}{5pt} % Adjust column spacing
	\renewcommand{\arraystretch}{0.9} % Reduce row height
	\caption{Balanced robustness results of ResNet-18 on CIFAR-10-LT (IR=10) under different hyperparameter settings.}
	\label{tab4}
	\begin{tabular}{cccccc}
		\noalign{\hrule height 1pt}  % 增加顶线粗细
		\\[-8pt] % 增加顶线与下一行之间的间距
		& & $\alpha = 0.1$ & $\alpha = 0.1$ & $\alpha = 0.05$ & $\alpha = 0.05$ \\
		& & $\gamma = 0.1$ & $\gamma = 0.05$ & $\gamma = 0.1$ & $\gamma = 0.05$ \\
		\noalign{\hrule height 0.5pt}  % 增加顶线粗细
		\\[-8pt] % 增加顶线与下一行之间的间距
		$\beta=0.1$ & & \textbf{35.15} & 34.7 & 33.14 & 31.24 \\
		$\beta=0.05$ & & 34.44 & 34.97 & 33.56 & 35.04 \\
		\bottomrule
	\end{tabular}
\end{table}
%\vspace{-8mm} % 减少两个表格之间的间距
\begin{table}[H]
	\centering
	\setlength{\tabcolsep}{4pt} % Reduce column spacing
	\renewcommand{\arraystretch}{1} % Reduce row height
	\caption{Balanced accuracy and balanced robustness using ResNet-18 on CIFAR-10-LT under different long-tail imbalance ratios(IRs). Better results are highlighted in\textbf{ bold}.}
	\label{tab5}
	\begin{tabular}{c c ccccc}
		\noalign{\hrule height 1pt}  % 增加顶线粗细
		\\[-8pt] % 增加顶线与下一行之间的间距
		\renewcommand{\arraystretch}{1.5} % 增加行高
		\textbf{IR} & \textbf{Method} & \textbf{Clean} & \textbf{FGSM} & \textbf{PGD-20} & \textbf{CW} & \textbf{AA} \\
		\midrule
		\multirow{2}{*}{10} & AT-BSL & 72.74 & 34.13 & 26.86 & 15.67 & 25.26 \\
		& TAET   & \textbf{74.67} & \textbf{39.59} & \textbf{35.15} & \textbf{57.59} & \textbf{30.57} \\
		\noalign{\hrule height 0.5pt}  % 增加顶线粗细
		\\[-8pt] % 增加顶线与下一行之间的间距
		\multirow{2}{*}{20} & AT-BSL & 66.4  & 28.82 & 22.56 & 15.43 & 21.46 \\
		& TAET   & \textbf{66.93} & \textbf{36.82} & \textbf{33.98} & \textbf{58.41} & \textbf{27.84} \\
		\noalign{\hrule height 0.5pt}  % 增加顶线粗细
		\\[-8pt] % 增加顶线与下一行之间的间距
		\multirow{2}{*}{50} & AT-BSL & 58.23 & 26.37 & 20.87 & 16.74 & 19.79 \\
		& TAET   & \textbf{59.37} & \textbf{31.69} & \textbf{29.35} & \textbf{50.15} & \textbf{25.3} \\
		\noalign{\hrule height 0.5pt}  % 增加顶线粗细
		\\[-8pt] % 增加顶线与下一行之间的间距
		\multirow{2}{*}{100} & AT-BSL & \textbf{54.73} & 22.51 & 19.17 & 17.34 & 18.29 \\
		& TAET   & 52.68 & \textbf{28.68} & \textbf{26.53} & \textbf{42.3} & \textbf{22.98} \\
		\bottomrule
	\end{tabular}
\end{table}
\noindent epochs increased clean sample accuracy by 7.19\% and balanced robustness under AA attack by 1.04\%, compared to omitting cross-entropy loss. This strategy also reduces computational costs, improving both model performance and training efficiency, as shown in Figure~\ref{fig:fig7}.

%-------------------------------------------------------------------------
\subsection{Ablation Study}
\label{sec:6.4}
We analyzed the proposed adversarial training on CIFAR-10-LT using ResNet-18\cite{resnet} for more insights:

\noindent \textbf{Contribution of HARL.} Fig.~\ref{fig:fig6b} shows the impact of each component. BCL notably improves accuracy, as expected. HDL and RCEL are designed to boost weak classes and perform best in combination.

\noindent \textbf{Hyperparameter Importance Analysis.} We examined the influence of hyperparameters $\alpha$, $\beta$, and $\gamma$ on performance, adjusting their ratios to assess stability. Due to squared terms in the loss, value fluctuations may impact training stability. Thus, we used a 40-epoch cross-entropy training period for stable observation. Tab.~\ref{tab4} details results across hyperparameter settings, showing that equal weights ($\alpha = \beta = \gamma$) yield the best performance. This emphasizes the importance of balanced hyperparameters in enhancing robustness and accuracy for long-tail datasets.
%-------------------------------------------------------------------------
\subsection{Results Across Long-Tail Imbalance Ratios}
\label{sec:6.5}

To evaluate the effectiveness of our method across different Imbalance Ratios (IR), we generated long-tail versions of CIFAR-10-LT with varying IRs. The results shown in Tab.~\ref{tab5} demonstrate that our method outperforms previous state-of-the-art techniques in terms of robustness under a variety of IR conditions. This enhanced robustness not only improves class adaptability but also boosts overall accuracy in imbalanced datasets, underscoring the practical potential of our approach for real-world applications.

% TODO: \usepackage{graphicx} required

\section{Conclusion}
\label{sec:Conclusion}

In this study, we analyze the limitations of AT-BSL in addressing long-tail robustness. Through extensive experiments, we show that the HARL strategy improves performance on both long-tail and weak classes. We propose balanced robustness as a novel metric to assess long-tail robustness, extending the concept of balanced accuracy. To mitigate robust overfitting, we introduce a two-stage adversarial equalization training (TAET) approach, which reduces overfitting while improving both adversarial robustness and clean accuracy. This method is computationally efficient and effective for practical applications. Our analysis confirms the generalizability and superiority of the proposed method. This work significantly advances adversarial training for real-world scenarios, enhancing adaptability to long-tail data. Future work will optimizing performance for long-tail distributions. Additionally, we will conduct experiments on real-world long-tail datasets to refine the model's adversarial robustness.
\newpage
\section*{Acknowledgments}
This work was supported in part by the Hefei Municipal Natural Science Foundation under Grant(No.HZR2403), Natural Science Research Project of Colleges and Universities in Anhui Province (No.2022AH051889, No.2308085QF227),  the Fundamental Research Funds for the Central Universities and the Fundamental Research Funds for the Central Universities of China.
{
	\small
	\bibliographystyle{ieeenat_fullname}
	\bibliography{main}
}

\clearpage
\setcounter{page}{1}
\maketitlesupplementary

\begin{appendix}
\section{Implementation Details of Experiments}
\label{sec:A}

\subsection{Training Details and Hyper-parameter Setting}
\label{sec:A.1}

We use ResNet-18 as the model architecture. The initial learning rate is set to 0.1, with decay factors of 10 at epochs 75 and 90, for a total of 100 epochs. We evaluate using the final epoch, and no methods employ early stopping.We use the SGD  optimizer with a momentum of 0.9 and weight decay set to 5e-4. In the main paper, we set the batch size to 128.For adversarial training, the maximum perturbation is set to 8/255, the step size to 2/255, and the number of steps to 10. To ensure fairness in the comparison, we adopt this same configuration for adversarial training across all methods.There are several hyperparameters involved, among which the most influential are  $\alpha$, $\beta$, $\gamma$, and the number of epochs during the initial phase of training. For CIFAR-10-LT, CIFAR-100-LT, and Dermamnist, we set  $\alpha$ = $\beta$ = $\gamma$ = 0.1 and use 40 epochs for training in the adversarial domain.

\subsection{Training Details and Hyper-parameter Setting}

For the defense methods compared in this paper, we use their official implementations, including AT-BSL\cite{changweiduikang2024}\footnote{\url{https://github.com/NISPLab/AT-BSL}}, RoBAL\cite{changweiduikang2021}\footnote{\url{https://github.com/wutong16/Adversarial Long-Tail}} , AT\cite{at1pgd}\footnote{\url{https://github.com/MadryLab/cifar10 challenge}}, TRADES\cite{trades}\footnote{\url{https://github.com/yaodongyu/TRADES}}, MART\cite{mart}\footnote{\url{https://github.com/NISPLab/AT-BSL}}, GAIRAT\cite{gairat}\footnote{\url{https://github.com/zjfheart/Geometry-aware-Instance-reweightedAdversarial-Training}}, LAS-AT\cite{lasat}\footnote{\url{7https://github.com/jiaxiaojunqaq/las-at}}, and REAT\cite{cahngweiduikang2023arxiv}\footnote{\url{https://github.com/GuanlinLee/REAT}}.
For the attacks used in the evaluation, we implement them based on the official code and the original papers, including FGSM\cite{duikang2fgsm}, PGD\cite{at1pgd}, CW\cite{cw}, and AutoAttack\cite{aa}.

\begin{figure}
	\centering
	\includegraphics[width=1\linewidth]{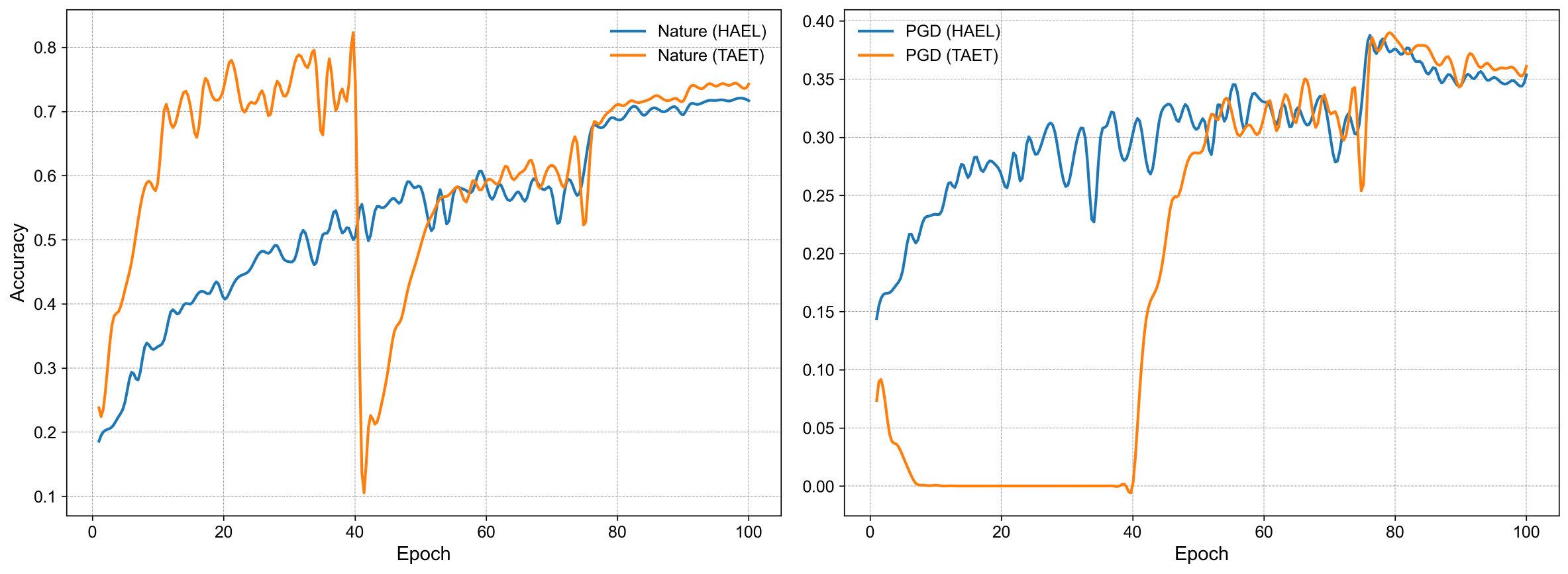}
	\caption{Variations in clean accuracy (left) and adversarial robustness (right) during training with the application or absence of two-stage training.}
	\label{fig:fig8}
\end{figure}

\section{Additional Experiments}

\subsection{Necessity and Benefits of the TAET Two-Stage Training}

The comparison between the two-stage training method (TAET) and the single-stage approach (HARL) is illustrated in the figure. TAET employs a two-stage training strategy, where the first 40 epochs are dedicated to optimizing cross-entropy (CE), followed by adversarial loss (HARL) optimization in subsequent epochs. This methodology enables a dynamic equilibrium between natural accuracy and adversarial robustness as shown in Figure~\ref{fig:fig8}.

The left graph compares TAET and HARL in terms of natural accuracy. During the initial 40 epochs of CE training, TAET primarily concentrates on improving the classification accuracy of natural samples. This phase enables rapid convergence and leads to a significant performance improvement over HARL, thus establishing a strong baseline for natural performance. After completing the 40th epoch, TAET transitions to HARL optimization. Although a transient fluctuation in natural accuracy occurs during this transition, the model quickly adapts to the new training objective and ultimately achieves a natural accuracy of approximately 0.8, clearly surpassing HARL. This outcome demonstrates that TAET's two-stage training approach provides sufficient room for optimizing adversarial robustness while maintaining high natural accuracy, all while offering a substantial reduction in resource requirements.

The right graph presents the comparison of adversarial robustness under PGD attacks. During the first 40 epochs of CE training, TAET predominantly focuses on optimizing natural accuracy, with minimal attention to adversarial robustness, resulting in nearly zero PGD robustness. However, following the switch to HARL optimization at epoch 40, TAET’s PGD robustness improves at an accelerated rate, eventually surpassing HARL in the later stages of training and stabilizing around 0.36. This finding underscores that TAET’s strategy effectively prioritizes natural accuracy optimization before shifting focus to adversarial robustness, achieving a dynamic balance between the two objectives. Additionally, by combining CE and HARL in a complementary manner, TAET not only significantly enhances the model's ability to generalize on natural samples but also demonstrates substantial potential for improving adversarial robustness, while significantly reducing the computational burden compared to HARL. These advantages collectively result in nearly halving the time and memory consumption, as shown in Figure~\ref{fig:fig6a}.

\begin{table}[t]
	\centering
	\caption{The standard accuracy performance of various methods ON CIFAR10-LT}
	\begin{tabular}{lcccccc}
		\toprule
		\textbf{Method} & \textbf{Clean} & \textbf{FGSM} & \textbf{PGD} & \textbf{CW} & \textbf{AA} \\
		\midrule
		AT-BSL & 77.74 & 43.56 & 36.53 & 22.77 & 34.99 \\
		TRADES & 75.66 & 39.17 & 32.71 & 40.76 & 31.83 \\
		AT & 77.96 & 44.98 & 37.7 & 24.41 & 36.09 \\
		MART & 69.8 & 44.45 & 39.66 & 34.31 & 38.24 \\
		ADT & 76.98 & 46.18 & 39.42 & 18.78 & 38.1 \\
		REAT & \textbf{79.72} & 41.25 & 33.39 & 17.5 & 31.85 \\
		LAS-AT & 76.02 & 44.16 & 38.27 & 51.73 & 36.15 \\
		HE & 77.61 & \textbf{46.18} & 38.73 & 22.3 & 37.26 \\
		GAIRAT & 71.49 & 41.77 & 36.87 & 54.6 & 33.79 \\
		\midrule
		TAET(ours) & 76.31 & 45.1 & \textbf{40.8} & \textbf{62.21} & \textbf{39.24} \\
		\bottomrule
	\end{tabular}
	\label{tab:cifar10lt_results}
\end{table}

\subsection{Conventional Evaluation Metrics on CIFAR10-LT}

The experimental results on CIFAR-10-LT using accuracy and robustness metrics are shown in Tab.~\ref{tab:cifar10lt_results}. From these results, it is evident that our method achieves high robustness while maintaining strong performance in terms of conventional accuracy, demonstrating the best performance across multiple adversarial attack scenarios, including PGD,CW, and AA. Additionally, we observe that the MART method exhibits robust performance under the long-tail distribution, but its accuracy on clean samples is relatively lower. This suggests that a promising avenue for further improvement could involve combining the MART method with long-tail recognition approaches to enhance robustness in long-tail settings while maintaining higher accuracy on natural samples.

\subsection{Different IRs}

We present the results on CIFAR-10-LT and CIFAR-100-LT datasets under different Imbalance Ratios (IR) in Tabs.~\ref{tab:cifar10_results},~\ref{tab:imbalanced_results}. The results show that our method outperforms the previous state-of-the-art method, AT-BSL, across all evaluation metrics, further confirming the effectiveness of our approach. Notably, in the standard accuracy evaluation on the CIFAR-10 dataset, our method slightly underperforms AT-BSL\cite{changweiduikang2024} on clean samples. This phenomenon can be attributed to the fact that our method places more emphasis on optimizing the performance of the tail classes, which have relatively fewer samples in the overall dataset. As a result, while the balanced accuracy might be higher, the standard accuracy could be lower. Nevertheless, even under the standard accuracy evaluation, our method still demonstrates superior adversarial robustness compared to previous methods, which further highlights the superiority of our approach.

% 第一个表格
\begin{table}[t]
	\centering
	\setlength{\tabcolsep}{4pt} % 减小列间距
	\caption{The standard accuracy performance under different Imbalance Ratios (IR) on CIRAR10-LT}
	\label{tab:cifar10_results}
	\begin{tabular}{c c ccccc}
		\noalign{\hrule height 1pt}  % 顶部线条加粗
		\\[-8pt] % 顶部线条与下一行之间的间距
		\textbf{IR} & \textbf{Method} & \textbf{Clean} & \textbf{FGSM} & \textbf{PGD} & \textbf{CW} & \textbf{AA} \\
		\midrule
		\multirow{2}{*}{10} & AT-BSL & \textbf{77.74} & 43.56 & 36.53 & 22.77 & 34.99 \\
		& TAET & 76.31 & \textbf{44.37} & \textbf{38.17} & \textbf{53.61} & \textbf{37.1} \\
		\noalign{\hrule height 0.5pt}  % 中间线条加粗
		\\[-8pt]
		\multirow{2}{*}{20} & AT-BSL & \textbf{78.3} & 45.4 & 37.96 & 27.99 & 36.4 \\
		& TAET & 77.57 & \textbf{47.66} & \textbf{41.69} & \textbf{55.51} & \textbf{39.9} \\
		\noalign{\hrule height 0.5pt}  % 中间线条加粗
		\\[-8pt]
		\multirow{2}{*}{50} & AT-BSL & \textbf{80.68} & \textbf{52.74} & 46.05 & 38.93 & 44.19 \\
		& TAET& 77.53 & 51.2 & \textbf{46.98} & \textbf{39.2} & \textbf{44.62} \\
		\noalign{\hrule height 0.5pt}  % 中间线条加粗
		\\[-8pt]
		\multirow{2}{*}{100} & AT-BSL & \textbf{81.92} & \textbf{54.52} & 49.03 & 43.66 & 45.78 \\
		& TAET & 74.09 & 51.98 & \textbf{49.95} & \textbf{66.71} & \textbf{47.78 }\\
		\bottomrule  % 底部线条加粗
	\end{tabular}
\end{table}

% 第二个表格
\begin{table}[t]
	\centering
	\setlength{\tabcolsep}{4pt} % 减小列间距
	\caption{The standard accuracy performance under Different Imbalance Ratios (IR) on CIFAR100-LT}
	\label{tab:imbalanced_results}
	\begin{tabular}{c c ccccc}
		\noalign{\hrule height 1pt}  % 顶部线条加粗
		\\[-8pt] % 顶部线条与下一行之间的间距
		\textbf{IR} & \textbf{Method} & \textbf{Clean} & \textbf{FGSM} & \textbf{PGD} & \textbf{CW} & \textbf{AA} \\
		\midrule
		\multirow{2}{*}{10} & AT-BSL & 48.41 & 24.77 & 20.81 & 18.8 & 18.71 \\
		& TAET & \textbf{55.69} & \textbf{30.2} & \textbf{24.46} & \textbf{29.69} & \textbf{22.25} \\
		\noalign{\hrule height 0.5pt}  % 中间线条加粗
		\\[-8pt]
		\multirow{2}{*}{20} & AT-BSL & 48.27 & 24.98 & 20.23 & 20.05 & 18.12 \\
		& TAET & \textbf{57.31} & \textbf{30.07} & \textbf{23.19} & \textbf{32.79} & \textbf{20.56} \\
		\noalign{\hrule height 0.5pt}  % 中间线条加粗
		\\[-8pt]
		\multirow{2}{*}{50} & AT-BSL & 49.46 & 26.19 & 22.99 & 24.43 & 20.35 \\
		& TAET & \textbf{60.71} & \textbf{30.17} & \textbf{24.05} & \textbf{35.7} & \textbf{21.59} \\
		\noalign{\hrule height 0.5pt}  % 中间线条加粗
		\\[-8pt]
		\multirow{2}{*}{100} & AT-BSL & 48.48 & 28.13 & 24.34 & 28.2 & 21.58 \\
		& TAET & \textbf{59.56} & \textbf{32.91} & \textbf{26.52} & \textbf{40.22} & \textbf{23.17} \\
		\bottomrule  % 底部线条加粗
	\end{tabular}
\end{table}

\begin{figure}
	\centering
	\includegraphics[width=0.8\linewidth]{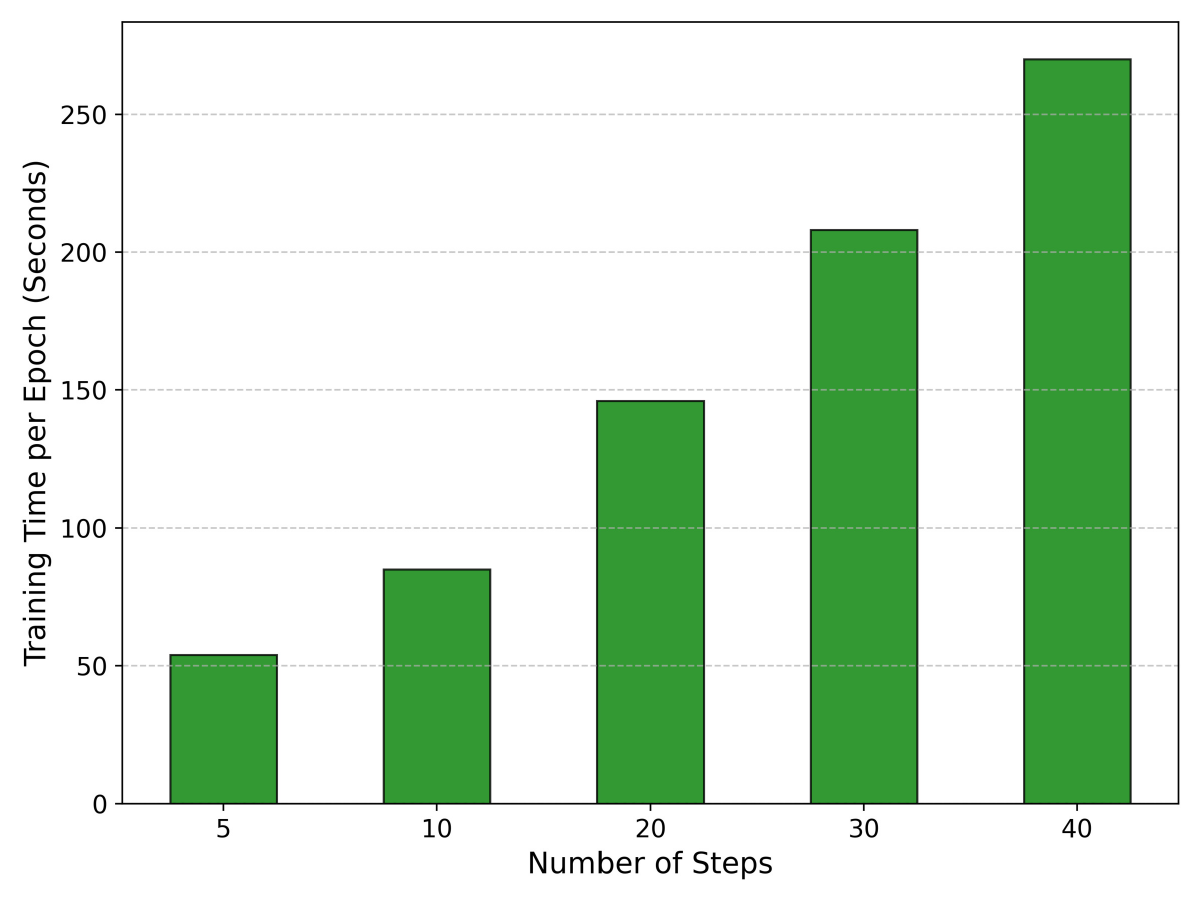}
	\caption{Time Cost Under Different Numbers of Steps}
	\label{fig:fig9}
\end{figure}

\subsection{Different PGD Steps }
Through an analysis of the model's accuracy on clean samples and its performance under various adversarial attacks across different training steps in Tabs.~\ref{tab:steps_impact}, along with an evaluation of the associated time cost, we observed that the number of steps significantly impacts both adversarial robustness and computational efficiency. Specifically, clean accuracy remained stable as the number of steps increased, consistently ranging from 73\% to 79\%, indicating that the model retains strong robustness in the absence of attacks. Under adversarial attacks, model accuracy improved with more steps, particularly under FGSM and CW attacks, peaking at 30 steps. However, the associated training time increased substantially, rising from 54 seconds at 5 steps to 208 seconds at 30 steps, and further to 270 seconds at 40 steps, which imposes significant computational overhead. Notably, using 10 steps provided a reasonable balance, achieving competitive adversarial robustness and clean accuracy while requiring only 85 seconds per epoch, which is significantly lower than the time cost for 30 or 40 steps. These results suggest that selecting 10 steps offers a practical compromise between adversarial robustness, clean accuracy, and computational efficiency.

\begin{table}[t]
	\centering
	\setlength{\tabcolsep}{8pt} % Adjust column spacing
	\renewcommand{\arraystretch}{1.2} % Adjust row height
	\caption{The Impact of Adversarial Training Steps on Adversarial Robustness}
	\label{tab:steps_impact}
	\begin{tabular}{c c c c c c}
		\noalign{\hrule height 1pt}  % 顶部线条加粗
		\textbf{Steps} & \textbf{Clean} & \textbf{FGSM} & \textbf{PGD} & \textbf{CW} & \textbf{AA} \\
		\midrule
		5  & 78.86 & 31.21 & 22.7  & 30.52 & 19.77 \\
		10 & 74.25 & 40.64 & 35.11 & 60.22 & 31.17 \\
		20 & 73.27 & 40.21 & 36.62 & 60.25 & 31.36 \\
		30 & 73.24 & 41.82 & 37.55 & 61.14 & 33.17 \\
		40 & 73.89 & 41.00 & 36.43 & 59.90 & 32.20 \\
		\bottomrule  % 底部线条加粗
	\end{tabular}
\end{table}

\subsection{Different PGD Size }
In Tabs.~\ref{tab:step_size_accuracy} we observed that step size adjustments significantly influence the model's adversarial robustness while having a relatively minor effect on clean sample performance. Specifically, clean accuracy remained stable across all step sizes, ranging between 73\% and 74\%, which highlights the model's robustness on natural samples. In contrast, adversarial robustness exhibited varying trends with step size changes. Larger step sizes, such as 3/255, notably improved robustness against stronger attacks like PGD and AA, with AA accuracy increasing from 31.17\% to 37.74\%. Meanwhile, robustness against FGSM attacks remained relatively stable, whereas performance under CW attacks showed some fluctuation as the step size increased. These findings suggest that increasing the step size can enhance robustness in certain adversarial scenarios but may also introduce vulnerabilities in others (e.g., CW attacks). Therefore, while appropriately increasing the step size can improve adversarial robustness without compromising clean accuracy, the optimal step size should be carefully selected based on task-specific requirements and the adversarial attack types encountered.
\begin{table}[ht]
	\centering
	\caption{The Impact of Adversarial Training Step Size on Adversarial Robustness}
	\begin{tabular}{cccccc}
		\toprule % 顶部线条加粗
		\textbf{Size} & \textbf{Clean} & \textbf{FGSM} & \textbf{PGD} & \textbf{CW} & \textbf{AA} \\
		\midrule % 中部线条
		1/255 & 74.25 & 40.64 & 35.11 & 60.22 & 31.17 \\
		2/255 & 73.13 & 39.88 & 35.60 & 58.26 & 31.05 \\
		3/255 & 73.29 & 40.58 & 36.20 & 58.56 & 37.74 \\
		\bottomrule % 底部线条加粗
	\end{tabular}
	\label{tab:step_size_accuracy}
\end{table}

\subsection{Different model }
In Tab.~\ref{tab:model_comparison}, we present a comparison between our method and the previous state-of-the-art method (AT-BSL) across different models. The results indicate that our method not only surpasses the original approach in terms of balanced accuracy but also in adversarial robustness, highlighting the superiority of our method. Furthermore, it can be observed that when training on the long-tailed dataset with ViT-b/16 (without using a pre-trained model), the AT-BSL method has completely failed, which demonstrates the effectiveness of our two-stage training approach in stabilizing model training.

\begin{table}[ht]
	\centering
	\caption{Comparison results with AT-BSL under different models on CIFAR-10-LT(IR = 10), with better results highlighted in \textbf{bold}.}
	\label{tab:model_comparison}
	\begin{tabular}{lccccc}
		\toprule
		\textbf{Model} & \textbf{Method} & \textbf{Clean} & \textbf{FGSM} & \textbf{PGD} &  \textbf{AA} \\
		\midrule
		\multirow{2}{*}{ResNet50} 
		& AT-BSL & 74.34 & 35.41 & 29.01  & 27.14 \\
		& TAET & \textbf{74.48} & \textbf{40.17} & \textbf{34.70}  & \textbf{31.27} \\
		\cmidrule{1-6}
		\multirow{2}{*}{ViT-B/16}
		& AT-BSL & 10.00 & 10.00 & 10.00  & 10.00 \\
		& TAET & \textbf{22.65} & \textbf{18.45} & \textbf{18.56} & \textbf{14.72} \\
		\bottomrule
	\end{tabular}
\end{table}

\section{Time Efficiency Analysis}
\subsection{Total Training Time Comparison}

The total time consumption for TAET (Two-Stage Adversarial Training) and standard AT (Adversarial Training) can be expressed as:
\begin{equation}
	\begin{aligned}
		T_{\text{TAET}} &= N_{\text{CE}} \cdot (F + B) + N_{\text{AT}} \cdot \left[ F + B + \kappa \cdot (F + B_{\text{adv}}) \right] \\
		T_{\text{AT}} &= N_{\text{total}} \cdot \left[ F + B + \kappa \cdot (F + B_{\text{adv}}) \right]
	\end{aligned}
\end{equation}
where:
\begin{itemize}
	\item $N_{\text{CE}} = 40$: Number of epochs for the Cross-Entropy (CE) stage.
	\item $N_{\text{AT}} = 60$: Number of epochs for the Adversarial Training (AT) stage.
	\item $N_{\text{total}} = 100$: Total number of training epochs.
\end{itemize}

\subsection{Acceleration Ratio Derivation}

The time acceleration factor is defined as:
\begin{equation}
	\eta = \frac{T_{\text{TAET}}}{T_{\text{AT}}} = \frac{N_{\text{CE}} \cdot \rho + N_{\text{AT}} \cdot (1 + \kappa \cdot \gamma)}{N_{\text{total}} \cdot (1 + \kappa \cdot \gamma)}
\end{equation}
where:
\begin{itemize}
	\item $\rho = \frac{F + B}{F + B + \kappa \cdot (F + B_{\text{adv}})} \approx 0.139$: Relative efficiency during the CE stage 
	\item $\gamma = \frac{F + B_{\text{adv}}}{F + B} \approx 0.619$: Adversarial training time expansion coefficient 
\end{itemize}

Substituting the parameters, we get:
\begin{equation}
	\eta \approx \frac{40 \times 0.139 + 60 \times (1 + 10 \times 0.619)}{100 \times (1 + 10 \times 0.619)}   \approx 0.608
\end{equation}
This indicates that TAET reduces the theoretical computational workload by approximately 39.2\% 

\section{Space Efficiency Analysis}

\subsection{Memory Requirement Model}

The peak memory requirement during training is given by:
\begin{equation}
	M = M_{\text{model}} + M_{\text{data}} + M_{\text{grad}} + \delta \cdot M_{\delta}
\end{equation}
where:
\begin{itemize}
	\item $M_{\delta} = b \cdot c \cdot h \cdot w \cdot d \approx 1629.42$ MB: Memory for adversarial components 
	\item $b = 128$: Batch size.
	\item $c = 3$: Number of channels.
	\item $h = w = 32$: Height and width of the input data.
	\item $d = 4$ bytes: Data type size.
	\item $\delta \in \{0, 1\}$: Adversarial training activation flag.
\end{itemize}

\subsection{Memory Requirement Comparison}

The memory difference between TAET and AT can be quantified as:
\begin{equation}
	\Delta M = M_{\text{TAET}} - M_{\text{AT}} =
	\begin{cases}
		-1629.42 \text{ MB} & \text{CE stage} \\
		0 & \text{AT stage}
	\end{cases}
\end{equation}

\subsection{Effective Memory Savings}

The memory saving efficiency is defined as:
\begin{equation}
	\xi = \frac{T_{\text{CE}}}{T_{\text{total}}} \cdot M_{\delta} = \frac{40}{100} \times 1629.42 \approx 651.77 \text{ MB}
\end{equation}
This indicates that TAET reduces the memory occupation by 1629.42 MB during 40\% of the training time.

\section{Algorithmic Framework for Two-Stage Adversarial Training}
The following pseudocode outlines the implementation of Two-Stage Adversarial Equalization Training (TAET). This algorithm effectively balances natural accuracy and adversarial robustness through staged optimization of cross-entropy and hierarchical adversarial loss. By initially focusing on natural accuracy and later shifting to robustness objectives, TAET ensures a dynamic equilibrium between performance and defense. Its efficient design also reduces computational overhead compared to single-stage approaches, making it a practical solution for adversarial training.

\begin{table}[h]
	\renewcommand{\arraystretch}{1.2} % 调整行距
	\centering
\resizebox{\linewidth}{!}{ % 缩放至单栏宽度
		\begin{tabular}{ll}
				\toprule
				\multicolumn{2}{l}{\textbf{Algorithm 1: Two-Stage Adversarial Equalization Training}} \\ % 添加标题行
				\midrule
				\multicolumn{2}{p{\linewidth}}{\textbf{Require:} Classifier $F$, clean samples $x_0$ and corresponding labels $y$, number of classes $K$, optimizer, step size $\eta$, perturbation size $\epsilon$, number of PGD attack steps $n$, total training epochs $T$, epochs for standard cross-entropy training $T_{\text{ce}}$, loss balance weights $\alpha$, $\beta$, $\gamma$} \\ \midrule
				1: & \textbf{for} $t = 1$ to $T_{\text{ce}}$ \textbf{do} \\ 
				2: & \quad Forward pass $x_0$ through $F$ \\
				3: & \quad Compute cross-entropy loss: \\
				4: & \quad $\mathcal{L}_{\text{ce}} = \text{CrossEntropyLoss}(F(x_0), y)$ \\
				5: & \quad Backpropagate and update $F$ to minimize $\mathcal{L}_{\text{ce}}$ \\
				6: & \textbf{End for} \\ 
				7: & \textbf{for} $t = T_{\text{ce}} + 1$ to $T$ \textbf{do} \\
				 & \quad Initialize adversarial sample: \\
				8: & \quad $x' = x_0 + 0.001 \cdot \mathcal{N}(0, 1)$ \\
				9: & \quad \textbf{for} $j = 1$ to $n$ \textbf{do} \\
				 & \qquad Compute adversarial loss on current adversarial sample: \\
				10: & \qquad $\mathcal{L}_{\text{ce}} = \text{CrossEntropyLoss}(F(x'), y)$ \\
				 & \qquad Update adversarial sample via gradient ascent: \\
				 & \qquad $\nabla x' = \text{grad}(\mathcal{L}_{\text{ce}}, x')$ \\
				11: & \qquad $x' = \text{Clip}(x' + \eta \cdot \text{sign}(\nabla x'), x_0 - \epsilon, x_0 + \epsilon)$ \\
				12: & \quad \textbf{End for} \\
				 & \quad Generate final adversarial example: \\
				13: & \quad \quad $x' = \text{Clip}(x', 0.0, 1.0)$ \\
				14: & \quad Forward pass adversarial sample $x'$ through $F$ \\
				 & \quad Compute Hierarchical Equalization Loss (HEL): \\
				15: & \quad $\mathcal{L}_{\text{HEL}} = \alpha \cdot \mathcal{L}_{\text{bal}} + \beta \cdot \mathcal{L}_{\text{hie}} + \gamma \cdot \mathcal{L}_{\text{rare}}$ \\
				16: & \quad Backpropagate and update $F$ to minimize $\mathcal{L}_{\text{HEL}}$ \\
				 17:& \textbf{End for} \\ \bottomrule
			\end{tabular}
	}

\end{table}

\section{Adversarial Attacks}

\noindent\textbf{The Fast Gradient Sign Method}  generates adversarial examples by perturbing the input data in the direction of the gradient of the loss function with respect to the input. The perturbation is controlled by a small step size \( \epsilon \). The formula for FGSM is given by:
	
	\begin{equation}
		x' = x + \epsilon \cdot \text{sign}(\nabla_x J(\theta, x, y))
	\end{equation}
	
	where \( x \) is the original input image, \( \epsilon \) is the perturbation magnitude, and \( \nabla_x J(\theta, x, y) \) is the gradient of the loss function \( J(\theta, x, y) \) with respect to the input \( x \), where \( \theta \) represents the model parameters and \( y \) is the true label. The adversarial example \( x' \) is created by adding the perturbation \( \epsilon \) to the input.

\noindent\textbf{Projected Gradient Descent (PGD)} attack is an iterative version of FGSM, where perturbations are applied multiple times in small steps. After each step, the perturbed image is projected back into a feasible region to ensure that the perturbations stay within a specified bound. The update rule for PGD is given by:
	
	\begin{equation}
		x^{t+1} = \Pi_{x + \mathcal{B}} \left( x^t + \alpha \cdot \text{sign}(\nabla_x J(\theta, x^t, y)) \right)
	\end{equation}
	
	where \( \alpha \) is the step size, \( \mathcal{B} \) is the set of allowed perturbations (typically the \( L_\infty \) ball), and \( \Pi \) denotes the projection operator that ensures the adversarial example stays within the allowed perturbation space.

\noindent\textbf{Carlini-Wagner (CW)} attack is an optimization-based method that seeks the minimal perturbation necessary to misclassify an input. It achieves this by minimizing a loss function that balances the perturbation size and the model's confidence in the incorrect class. The objective function for the CW attack is:

\begin{equation}
	\begin{aligned}
		\min_{\delta} & \quad \| \delta \|_2 + c \cdot f(x + \delta) \\
		\text{s.t.} & \quad x + \delta \in [0,1]^n
	\end{aligned}
\end{equation}

Here, \( \delta \) is the perturbation added to the original input \( x \), \( \| \delta \|_2 \) denotes the \( L_2 \) norm of the perturbation, and \( c \) is a constant that balances the importance of the perturbation size against the attack success. The function \( f(x + \delta) \) is defined as:

\begin{equation}
	f(x + \delta) = \max \left( \max_{i \neq t} \{ Z(x + \delta)_i \} - Z(x + \delta)_t, -\kappa \right)
\end{equation}

In this function, \( Z(x + \delta)_i \) represents the logits (pre-softmax outputs) of the model for class \( i \), \( t \) is the target class in a targeted attack, and \( \kappa \) (kappa) is a confidence parameter that controls the margin by which the adversarial example's logit for the target class must exceed that of any other class. The constraint \( x + \delta \in [0,1]^n \) ensures that the perturbed input remains within valid data bounds. The goal is to find the smallest perturbation \( \delta \) that causes the model to misclassify the input as the target class \( t \) with high confidence.
	
\noindent\textbf{AutoAttack (AA)} is a state-of-the-art adversarial attack method that combines multiple attack strategies to generate robust adversarial examples. AA includes attack methods such as APGD, APGDT, FAB, and Square. Each attack method in AutoAttack is designed to explore different adversarial strategies, making it a versatile and powerful tool for generating adversarial examples across a wide range of models.

\end{appendix}

\end{document}